\documentclass[runningheads]{llncs}
\usepackage{graphicx}
\usepackage{amsmath,amssymb} % define this before the line numbering.
\usepackage{color}
\usepackage{colortbl}
\usepackage{epsfig}
\usepackage{multirow}
\usepackage{algorithm}
\usepackage{algpseudocode}
\usepackage{tabularx}
\usepackage{booktabs}

\usepackage[breaklinks=true,bookmarks=false]{hyperref}

\newcommand{\vect}[1]{\boldsymbol{#1}}
\def\eg{\emph{e.g}.}
\def\ie{\emph{i.e}.}
\def\wrt{{w.r.t}.~}

\def\rot#1{\rotatebox{90}{#1}}
\definecolor{tabrow}{rgb}{0.88,0.88,0.88}
\definecolor{Gray}{gray}{0.8}

\begin{document}

%macro for raising the point in decimal numbers; see example in the abstract
\newcommand{\point}{
    \raise0.7ex\hbox{.}
    }

%Do   -- NOT --    use any additional macros

\pagestyle{headings}

\mainmatter

%===========================================================
\title{Variational Gaussian Process Auto-Encoder for Ordinal Prediction of Facial Action Units} % Replace with your title

\titlerunning{Variational Gaussian Process Auto-Encoder} % Replace with your title

\authorrunning{Eleftheriadis {\em et al}.} % Replace with your names

\author{\!Stefanos Eleftheriadis$^*$ Ognjen Rudovic$^*$ Marc Peter Deisenroth$^*$ Maja Pantic$^{*\dagger}$\\
{\small \{\href{mailto:stefanos@imperial.ac.uk}{stefanos}, \href{mailto:orudovic@imperial.ac.uk}{orudovic}, \href{mailto:m.deisenroth@imperial.ac.uk}{m.deisenroth}, \href{mailto:m.pantic@imperial.ac.uk}{m.pantic}\}@imperial.ac.uk}} % Replace with your names
\institute{$^*$Department of Computing, Imperial College London, UK\\
$^\dagger$EEMCS, University of Twente, The Netherlands\\} % Replace with your institute's address

\maketitle

%===========================================================
\begin{abstract}
We address the task of simultaneous feature fusion and modeling of discrete ordinal outputs. We propose a novel Gaussian process (GP) auto-encoder modeling approach. In particular, we introduce GP {\em encoders} to project multiple observed features onto a latent space, while GP {\em decoders} are responsible for reconstructing the original features. Inference is performed in a novel variational framework, where the recovered latent representations are further constrained by the ordinal output labels. In this way, we seamlessly integrate the ordinal structure in the learned manifold, while attaining robust fusion of the input features. We demonstrate the representation abilities of our model on benchmark datasets from machine learning and affect analysis. We further evaluate the model on the tasks of feature fusion and joint ordinal prediction of facial action units. Our experiments demonstrate the benefits of the proposed approach compared to the state of the art.
\end{abstract}

%===========================================================
\section{Introduction}

Automated analysis of facial expressions has attracted significant attention because of its practical importance in psychology studies, human-computer interfaces, marketing research, and entertainment, among others~\cite{bartlett2010automated}. The most objective way to describe facial expressions is by means of the facial action coding system (FACS)~\cite{ekman2002facial}. This is the most comprehensive anatomically-based system that can be used to describe virtually all possible facial expressions in terms of 30+ facial muscle movements, named action units (AUs). FACS also defines rules for scoring the intensity of each AU in the range from absent to maximal intensity on a six-point ordinal scale. Therefore, FACS is critical for high-level interpretation of facial expressions. For instance, the high intensity of AU12 (lip corner puller), as in full-blown smiles, may indicate joy. Conversely, its low intensity may indicate fake smiles as in the case of sarcasm.

The machine analysis of AU intensities is challenging mainly due to the complexity and subtlety of human facial behavior as well as individual differences in expressiveness and variations in head-pose, illumination, occlusions, etc.~\cite{pantic2009machine}. These sources of variation are typically accounted for at the feature level by means of geometric- and appearance-based features, capturing the geometry and texture changes in a face, respectively. Furthermore, some AUs usually appear in combination with other AUs. For instance, the criteria for intensity scoring of AU7 (lid tightener) are changed significantly if AU7 appears with a maximal intensity of AU43 (eye closure) since this combination changes the appearance as well as timing of these AUs~\cite{rudovic2015context}. Furthermore,
co-occurring AUs can be non-additive, \eg, if one AU masks another a new and distinct set of appearances is created~\cite{ekman2002facial}. Thus, combining different facial features while  accounting for AU co-occurrences in a common framework is expected to result in a robust and more accurate estimation of target AUs intensity.

Most existing approaches to AU intensity estimation model each AU independently and cast it as a classification~\cite{rudovic2015context,mahoor2009framework,mavadati2013disfa,ming2015facial,valstar2015fera} or regression~\cite{savran2012regression,kaltwang2012continuous,jeni2013continuous,kaltwang2015doubly} task. While classification seems to be a natural choice to handle the problem, the related literature fails to account for the ordinal nature of the target intensity levels (misclassification of different levels is equally penalized). The regression-based approaches model the intensity levels on a continuous scale, which is sub-optimal when dealing with discrete outputs. Similarly, the models that do attempt multiple AU intenisty estimation (\eg,~\cite{li2013unified,sandbach2013markov,kaltwang2015latent,nicolle2015facial,mohammadi2016intensity}) adopt the same sub-optimal approach to deal with the nature of the output as the independent methods. However, they have showed improved performance in the target task due to the  modeling of AU co-occurrences. Apart from a few exceptions that treat each AU independently~\cite{savran2012regression,kaltwang2012continuous,ming2015facial}, none of the aforementioned approaches addresses the task of joint output modeling (\ie, multiple AUs)  while accounting for different modalities in the input (\ie, fusion of geometric and appearance features). These limitations can naturally be addressed by following recent advances in manifold learning~\cite{damianou2012manifold,urtasun2008transferring,Calandra2016} and, in particular,  using the framework of Gaussian processes (GPs)~\cite{rasmussen2006gaussian}. Within this framework, the problem of feature fusion is transformed to that of learning from multiple views, while continuous-valued predictions can be handled efficiently, for more than one output. However, as with the  regression-based models described above, these models treat the ordinal labels as continuous values. This also limits their potential to unravel an `ordinal' manifold, needed to facilitate estimation of target ordinal intensities.

In this work, we propose a novel manifold-based GP approach based on the Bayesian GP latent variable model (B-GPLVM)~\cite{titsias2010bayesian} that performs simultaneously the feature fusion and joint estimation of the AU ordinal intensity. Specifically, we propse the variational GP auto-encoder (VGP-AE), which is composed of a probabilistic {\em recognition} model, used to project  the observed features onto the manifold, and a generative model, used for their reconstruction. This, in contrast to existing work (\eg,~\cite{dai2015variational}) that applies deterministic  back-mappings, allows us to explicitly model the uncertainty in the projections onto the learned manifold. Additionally, we  endow the proposed VGP-AE with the ordinal outputs~\cite{agresti2010analysis}. The fusion of the information from the input features and learning of the joint ordinal output is performed simultaneously in a joint Bayesian framework. In this way, we seamlessly integrate the ordinal structure into the recovered manifold while attaining robust fusion of the target features. To the best of our knowledge, this is the first approach that achieves simultaneous feature fusion and joint AU intensity estimation in the context of facial behavior analysis.

\section{Related Work on AU Intensity Estimation}
To date, most existing work on automated analysis focuses on the detection of AU activations~\cite{mahoor2011facial,chu2013selective,zhao2015joint,eleftheriadis2015multi}. The problem of AU intensity estimation is relatively new in the field. Most of the research in this area focuses on independent modeling of AU intensities~\cite{rudovic2015context,mahoor2009framework,mavadati2013disfa,ming2015facial,valstar2015fera,savran2012regression,kaltwang2012continuous,jeni2013continuous,kaltwang2015doubly}. Only recently,  joint estimation of the intensity levels has been addressed~\cite{li2013unified,sandbach2013markov,kaltwang2015latent,nicolle2015facial,mohammadi2016intensity}. This is motivated by the fact that intensity annotations are difficult to obtain (due to the tedious process of manually coding) and that AU levels are highly imbalanced. Thus, by imposing the structure on the output in terms of AU co-occurrences robust intensity estimation is expected. 

Toward this direction,~\cite{li2013unified} proposed a two-stage learning strategy, where a multi-class support vector machine (SVM) is first trained for each AU independently. Then, the structure modeling is handled via a dynamic Bayesian network, which captures the semantic relationship among the AU-specific SVMs. In a similar fashion,~\cite{sandbach2013markov} used support vector regressors (SVR)  and a Markov random field (MRF). However, these two-stage approaches are sub-optimal for the target task as the regressors/classifiers and the AU relations are learned independently. %Thus, the structure of the features cannot explicitly affect the learning of the labels' dependencies, and vice versa, a fact which may result in loss of information. Moreover, both~\cite{li2013unified,sandbach2013markov} use information only from appearance features, which makes them more susceptible to subject and illumination variations. 
To overcome this,~\cite{kaltwang2015latent} proposed to learn latent trees that encode both the input features and (multiple) output AU labels. The structure of the latent variables is modeled using a tree-like graph. However, in the presence of high-dimensional inputs and multiple AUs, this method becomes prohibitively expensive. Moreover, the authors show that with this approach the fusion of different features does not benefit the estimation of AU intensity, achieving similar performance to when individual modalities are used. More recently,~\cite{mohammadi2016intensity} proposed a sparse learning approach that uses the notion of robust principal component analysis~\cite{candes2011robust} to decompose expression from facial identity. Then, joint intensity estimation of multiple AUs is performed via a regression model based on dictionary learning. However, this approach can deal with a single modality only. \cite{nicolle2015facial} casts the joint AU intensity estimation as a multi-task learning problem based on kernel regression (MLKR). However, in their formulation of the model, the use of MLKR does not scale to high-dimensional features, let alone when using features of different modalities  (\eg, geometric and appearance). 

The work presented in this paper advances the current state of the art in several aspects: (1) The proposed VGP-AE can efficiently perform the fusion of multiple modalities by means of a shared manifold; (2) Automatic feature selection is implicitly performed via the manifold. The recovered latent representations are used as input to multiple {\em ordinal} regressors~\cite{agresti2010analysis}, which are concurrently learned in a joint Bayesian framework; (3) GPs allow us to efficiently deal with high-dimensional input and output variables without significantly affecting the model's complexity.

\section{Variational Gaussian Process Auto-Encoder}
We assume that we have access to a training data set $\mathcal{D} = \lbrace \vect{Y}, \vect{Z}\rbrace$, which is comprised of $V$ observed input channels $\vect{Y} = \{\vect{Y}^{(v)}\}_{v = 1}^V$, and the associated output labels $\vect{Z}$.
Each input channel consists of $N$ i.i.d. samples $\vect{Y}^{(v)} = \{\vect{y}^{(v)}_i\}_{i=1}^{N}$, where $\vect{y}^{(v)}_i \in \mathbb{R}^{D_v}$ denotes corresponding facial features. $\vect{Z} = \{\vect{z}_i\}_{i=1}^{N}$ is the common label representation, where $z_{ic} \in \{1,\dots,S\}$ denotes the discrete, ordinal state of the $c$-th output (\ie, AU intensity level), $c = 1,\dots,C$. We are interested in simultaneously addressing the tasks of feature fusion and ordinal prediction of the multiple outputs.
For this purpose, we propose an approach that resembles recent work of generative models~\cite{kingma2013auto,rezende2014stochastic}. In these models, auto-encoders are employed to learn compact representations of the input data. In a standard auto-encoding setting, the encoding/decoding functions are modeled via neural networks. Here we replace these functions with probabilistic non-parametric mappings, significantly reducing the number of optimized parameters, and naturally modeling the uncertainty in the mappings. The proposed approach can be regarded as a B-GPLVM (generative model) with a fast inference mechanism based on the non-parametric, probabilistic mapping (recognition model). To achieve this, we impose GP priors on both models, and hence, obtain a well-defined GP-{\em encoder}, in accordance to the GP-{\em decoder}.

%$\vect{Y}^{(v)}$ represents different types of corresponding facial features and $\vect{Z}$ the multiple AU intensity levels.
%Each observed input space consists of $N$ i.i.d samples $\vect{Y}^{(v)} = \{\vect{y}^{(v)}_i\}_{i=1}^{N}$, where $\vect{y}^{(v)}_i \in \mathbb{R}^{D_v}$. 
%In what follows, we present our fully probabilistic auto-encoder for the framework of GPs.
\subsection{The Model}
\begin{figure}[t]
\centering
\footnotesize
\setlength{\tabcolsep}{2pt}
\begin{tabular}{ccc}
\includegraphics[scale=.64]{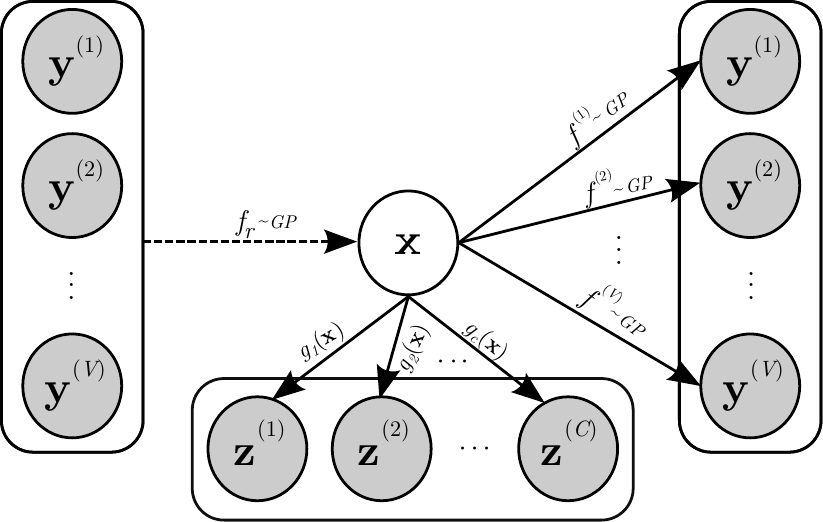} &
\raisebox{20pt}{\includegraphics[scale=.64]{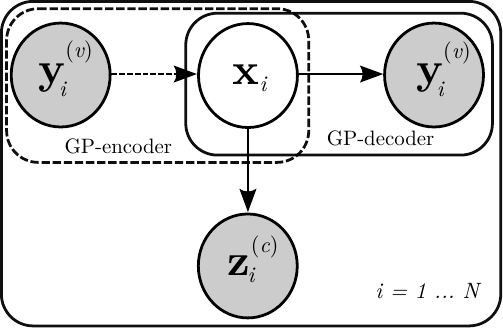}} & 
\raisebox{30pt}{\includegraphics[scale=.64]{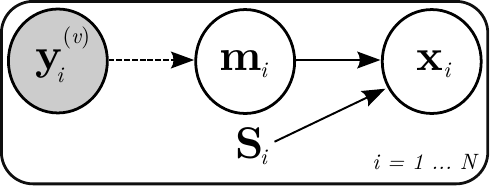}}\\
(a) Graphical model of VGP-AE & (b) Plate diagram & (c) Recognition model
\end{tabular}
\caption{{\footnotesize The proposed VGP-AE. (a) 
$f^{(v)}$ and $f_r$ are the GP-decoder and GP-encoder, respectively.
 The projection of the latent variable $\vect{x}$ to the labels' ordinal plane is facilitated through the ordinal regression $g(\vect{x})$. (b) Compact representation of the model. (c) The proposed recognition model (GP-encoder) with the intermediate variable $\vect{m}$.}}
\label{fig_model}
\end{figure}

Within the above setting, we assume that the observed features $\vect{Y}^{(v)}$ are generated by a random process, involving a latent (unobserved) set of variables $\vect{X} = \{\vect{x}_i\}_{i=1}^{N}, \vect{x}_i \in \mathbb{R}^{q}$, with $q\ll D_v$. The data pairs $\mathcal{D} = \lbrace \vect{Y}, \vect{Z}\rbrace$ are assumed to be conditionally independent given the latent variables, \ie, $\vect{Y} \perp\!\!\!\perp \vect{Z} | \vect{X}$. The random process of recovering the latent variables has two distinctive stages: (a) a latent variable $\vect{x}_i$ is generated from some general prior distribution $p(\vect{x})=\mathcal{N}(\vect{0}, \vect{I})$, and further projected to the labels' ordinal plane via $p(\vect{z}|\vect{x})$; (b) an observed input $\vect{y}^{(v)}_i$ is generated from the conditional distribution $p(\vect{y}^{(v)}|\vect{x})$. This process is described in Fig.~\ref{fig_model}(a),(b). Using this approach, we can now perform classification in the lower-dimensional space of $\vect{X}$. However, this requires access to the intractable true posterior $p(\vect{x}|\vect{y}^{(v)})$. %, which in the B-GPVLM is approximated by a variational distribution $q(\vect{x})$~\cite{titsias2010bayesian}.
%$(\vect{y}^{(v)}_i, \vect{z}_i) \perp\!\!\!\perp (\vect{y}^{(v)}_j, \vect{z}_j) | \vect{x}_i, i\neq j$
%, since the dimensionality of the latent variables is much smaller than that of the observations

To constrain the distribution of the latent variables we follow~\cite{kingma2013auto,rezende2014stochastic} and introduce the {\em recognition} model $p_r(\vect{x}|\vect{y}^{(v)})$. Hence, we end up with a supervised auto-encoder setting
\begin{equation}\label{ae}
\small
 \vect{y}_{i}^{(v)}|\vect{x}_{i} = f^{(v)}(\vect{x}_{i}; \vect{\theta}^{(v)}) + \epsilon^{(v)},  \quad \vect{x}_{i}|\vect{y}_{i}^{(v)} = f_r(\vect{y}_{i}^{(v)}; \vect{\theta}_r) + \epsilon_r, \quad \vect{z}_{i}|\vect{x}_{i} = g(\vect{x}_{i};\vect{W}),
\end{equation}
where the latent space is further encouraged to reflect the structure of the output labels.
Here, $\epsilon^{(v)}\sim\mathcal{N}(\vect 0,\sigma_v^2\vect I)$, $\epsilon_r\sim\mathcal{N}(\vect 0,\sigma_r^2\vect I)$. We place GP priors on $f^{(v)},f_r$ with corresponding hyper-parameters $\vect{\theta}^{(v)}, \vect{\theta}_r$.\footnote{The subscript $r$ indicates that the process facilitates the recognition model.} $g$ denotes the ordinal regression that transforms the latent variables to the labels' ordinal plane, via $\vect{W} = \{ \vect{w}_c\}_{c=1}^C, \vect{w}_c\in \mathbb{R}^q$.

In the following, we detail how to learn the GP auto-encoder in Eq.~(\ref{ae}) by deriving a variational approximation to the log-marginal likelihood 
\begin{equation}\label{marg_lklhd}
\log p(\vect{Y},\vect{Z}) = \log\int p(\vect{Z}|\vect{X})\prod\nolimits_v p(\vect{Y}^{(v)}|\vect{X}) p(\vect{X}) d\vect{X}.
\end{equation}

\subsection{Deriving the Lower Bound}\label{sec:vae}

We exploit the conditional independence property of $\vect{Y} \perp\!\!\!\perp \vect{Z} | \vect{X}$ and focus our analysis on the GP auto-encoder. The ordinal information from the labels is incorporated in the presented variational framework in Sec.~\ref{sec:ord}. As in~\cite{eleftheriadis2015multi}, we place GP priors on $f^{(v)},f_r$, and after integrating out the mapping functions, we obtain the conditionals
\begin{equation}
 p(\vect{Y}^{(v)}|\vect{X}) = \mathcal{N}(\vect{0}, \vect{K}^{(v)} + \sigma_v^2\vect{I}),  \qquad  p_r(\vect{X}|\vect{Y}) = \mathcal{N}(\vect{0}, \vect{K}_r + \sigma_r^2\vect{I}),
\end{equation}
where {\small $\vect{K}^{(v)} = k^{(v)}(\vect{X},\vect{X})$} and {\small $\vect{K}_r = \sum_v k_r^{(v)}(\vect{Y}^{(v)},\vect{Y}^{(v)})$} are the kernels associated with each process. Note that in the recognition model the relevant kernel allows us to easily  combine  multiple  features via the sum of the individual kernel functions. Training of the recognition model consists of maximizing the conditional $p_r(\vect{X}|\vect{Y})$ \wrt the kernel hyper-parameters $\vect{\theta}_r$. For the generative model we maximize the marginal likelihood (labels $\vect{Z}$ are omitted here)
\begin{equation}\label{likelihood}
p(\vect{Y}) = \int \prod\nolimits_{v=1}^V p(\vect{Y}^{(v)}|\vect{X})p(\vect{X})d\vect{X}.
\end{equation}
Since the above integral is intractable, we resort to approximations.
Our main interest is to recover a Bayesian non-parametric solution for both the GP encoder and decoder. We first need to break the circular dependence between $\vect{Y}^{(v)}$ and $\vect{X}$ in order to train the two GPs simultaneously. 

\paragraph{GP-encoder.} We decouple $\vect{X}$ and $\vect{Y}$ by introducing an intermediate variable $\vect{M}=\{ \vect{m}_i\}_{i=1}^N$, so that the recognition model becomes $\vect{y}^{(v)} \rightarrow \vect{m} \rightarrow \vect{x}$. The GP operates on $\vect{y}^{(v)}, \vect{m}$, while $\vect{x}$ is the noisy observations of $\vect{m}$. This process is described in Fig.~\ref{fig_model}(c). We follow a mean field approximation and introduce the variational distribution $q(\vect{X}|\vect{M}) = \prod_i q_i(\vect{x}_i|\vect{m}_i) = \prod_i \mathcal{N}(\vect{m}_i, \vect{S}_i)$. Here, $\vect{m}_i, \vect{S}_i \in \mathbb{R}^q$ are variational parameters\footnote{For simplicity we assume an isotropic (diagonal) covariance across the dimensions.} of $q_i$. %, while $\vect{m}_i$ follows the distribution $p(\vect{M}|\vect{Y})$.
We define $\vect{M}$ by employing the cavity distribution of the leave-one-out solution of GP~\cite{rasmussen2006gaussian} 
%We define the latter by employing the {\em cavity} distribution of the leave-one-out solution of GP~\cite{rasmussen2006gaussian} so that
% facilitates the intermediate layer between $\vect{y}^{(v)}$ and $\vect{x}$. Hence, the recognition model now utilizes the GP mapping from $\vect{y}^{(v)}$ to $\vect{m}$, \ie, variational mean of $q(\vect{x})$. To define this distriution, and in order to retain the factorization across the $N$ datapoints
\begin{equation}\label{cavity}
p(\vect{M}|\vect{Y}) = \prod\nolimits_i p(\vect{m}_i|\vect{Y}, \vect{M}_{\backslash i}) =  \prod\nolimits_i \mathcal{N}(\hat{\vect{m}}_i, \hat{\sigma}_i^2\vect{I}),
\end{equation}
where the subscript $\backslash i$ means `all datapoints except $i$', and the mean and variance of the Gaussian are given by~\cite{rasmussen2006gaussian}
\begin{equation}\label{loo_mv}
\hat{\vect{m}}_i = \vect{m}_i - \left[\vect{K}_r^{-1}\vect{M} \right]_i / \left[\vect{K}_r^{-1} \right]_{ii}, \qquad \hat{\sigma}_i^2 = 1 / \left[\vect{K}_r^{-1} \right]_{ii}.
\end{equation}
%Eq.~(\ref{cavity})--\eqref{loo_mv} reveals that $\vect{m}$ is no longer a free-form variational parameter, but it is a random variable.\todo{Is this important? Can't we just say that we marginalize it out?} 
We now integrate out the intermediate layer and propagate the uncertainty of the GP mapping to the latent variable $\vect{X}$, which yields the variational distribution
\begin{equation}\label{var_dist}
q(\vect{X}|\vect{Y}) = \prod\nolimits_i \mathcal{N}(\hat{\vect{m}}_i, \vect{S}_i + \hat{\sigma}_i^2\vect{I}).
\end{equation}
\paragraph{GP-decoder.} The proposed recognition model, \ie, the variational distribution of Eq.~(\ref{var_dist}), can be employed to approximate the intractable marginal likelihood of Eq.~(\ref{likelihood}). By introducing the variational distribution as an approximation to the true posterior, and after applying the Jensen's inequality, we obtain the lower bound to the log-marginal likelihood (again, labels $\vect{Z}$ are omitted)
\begin{equation}\label{elbo}
\log p(\vect{Y}) \ge \mathcal{F}_1 = \sum\nolimits_v \mathbb{E}_{q(\vect{X}|\vect{Y})}
\left[ \log p(\vect{Y}^{(v)}|\vect{X})\right] - KL(q(\vect{X}|\vect{Y})||p(\vect{X})).
\end{equation}
Training our model consists of maximizing the lower bound of Eq.~(\ref{elbo}) \wrt the variational parameters $\vect{M},\vect{S}$ and the hyper-parameters of the kernels $\vect{K}^{(v)},\vect{K}_r$. Further details are given in Sec.~\ref{sec_optim}.

\subsection{Incorporating Ordinal Variables}\label{sec:ord}
In the previous section, we presented the recognition model that we employ to learn a nonlinear manifold from the observed inputs. In the following, we further constrain this manifold by imposing an ordinal structure. This is attained by introducing ordinal variables that account for $C$ ordinal levels of AUs. We use the notion of ordinal regression~\cite{agresti2010analysis} and, in particular, the ordinal threshold model that imposes the monotonically increasing structure of the discrete output labels to the continuous manifold.
%monotonically increasing constraints on the intensity levels of each  output.
 Formally, the non-linear mapping between the manifold $\vect{X}$ and the ordinal outputs $\vect{Z}$ is modeled as
\begin{equation}\label{ord_noisy}
\small
\!\!\!\!p(\vect{Z}|g(\vect{X})) = \prod_{i,c} p(z_{ic}|g_c(\vect{x}_i)), \quad\!\!\!
p(z_{ic} = s|g_c(\vect{x}_i)) = 
\begin{cases}
1 \quad $if $ g_c(\vect{x}_i) \in (\gamma_{c,s-1}, \gamma_{c,s}]\\
0 \quad $otherwise$,
\end{cases}
\end{equation}
where $i=1,\dots,N$ indexes the training data. $\gamma_{c,0} = -\infty \le \cdots \le \gamma_{c,S} = +\infty$ are the thresholds or cut-off points that partition the real line into $s=1,\dots,S$ contiguous intervals. 
%and $\gamma_{c,s}=\gamma_{c,1}+ \sum_{t=2}^{s-1}   \triangle _{s,t}$, where $\triangle _{s,t}$ are positive padding variables and $t=2,\dots,S-1$.
 These intervals map the real function value $g_c(\vect{x})$ into the discrete variable $s$, corresponding to each of $S$ intensity levels of an AU, while enforcing the ordinal constraints. The threshold model $p(z_{ic} = s|g_c(\vect{x}_i))$ is used for ideally noise-free cases. Here, we assume that the latent functions $g_c(\cdot)$\footnote{Note that we adopt here a linear model for $g_c(\cdot)$ as it operates on a low-dimensional non-linear manifold $\vect{X}$, already obtained by the GP auto-encoder.} are corrupted by Gaussian noise, leading to the following formulation
\begin{equation}
g_c(\vect{x}_i) = \vect{w}_c^T\vect{x}_i + \epsilon_g, \quad \epsilon_g\sim\mathcal{N}(0,\sigma_g^2).
\end{equation}
By integrating out the noisy projections from Eq.~(\ref{ord_noisy}) (see~\cite{chu2005gaussian} for details), we arrive at the ordinal log-likelihood
\begin{equation}\label{ord_lklhd}
\small
\log p(\vect{Z} | \vect{X}, \vect{W}) = \sum\nolimits_{i,c}\mathbb{I}(z_{ic}=s) \log\left( \Phi\left(\frac{\gamma_{c,s} - \vect{w}_c^T\vect{x}_i}{\sigma_g}\right) - 
\Phi\left(\frac{\gamma_{c,s-1} - \vect{w}_c^T\vect{x}_i}{\sigma_g}\right)\right),
\end{equation}
where $\Phi(\cdot)$ is the Gaussian cumulative density function, and $\mathbb{I}(\cdot)$ is the indicator function. Finally, by using the ordinal likelihood defined in Eq.~(\ref{ord_lklhd}), we obtain the final lower bound of our log-marginal likelihood
\begin{small}
\begin{align}\label{final_elbo}
\nonumber\log &p(\vect{Y},\vect{Z} | \vect{W}) \ge \mathcal{F}_2 = 
\sum\nolimits_v \mathbb{E}_{q(\vect{X}|\vect{Y})}
\big[ \log p(\vect{Y}^{(v)}|\vect{X})\big] - KL(q(\vect{X}|\vect{Y})||p(\vect{X}))\\
&+\sum_{i,c}\mathbb{I}(z_{ic}=s) \mathbb{E}_{q(\vect{X}|\vect{Y})}
\left[\log\left( \Phi\left(\frac{\gamma_{c,s} - \vect{w}_c^T\vect{x}_i}{\sigma_g}\right) - 
\Phi\left(\frac{\gamma_{c,s-1} - \vect{w}_c^T\vect{x}_i}{\sigma_g}\right)\right)\right].
\end{align}
\end{small}%
%
%Note that we assume here that the noise component is independent for each of $C$ ordinal outputs; however, their dependencies are captured through the shared manifold $\vect{X}$.

\subsection{Learning and Inference}\label{sec_optim}
Training our model consists of maximizing the lower bound of Eq.~(\ref{final_elbo}) \wrt the variational parameters $\{\vect{S},\vect{M}\}$, the hyper-parameters $\{\vect{\theta}^{(v)},\sigma_v,\vect{\theta}_r^{(v)},\sigma_r\}$ of the GP mappings, and the parameters $\{ \vect{W}, \gamma, \sigma_g \}$ of the ordinal classifier. For the kernel of the GP-decoder we use the radial basis function (RBF) with automatic relevance determination (ARD), which can effectively estimate the dimensionality of the latent space~\cite{damianou2012manifold}. For the kernel of the GP-encoder we use the isotropic RBF for each observed input.
To utilize a joint optimization scheme, we use stochastic backpropagation~\cite{kingma2013auto,rezende2014stochastic}, where the re-parameterization trick is applied in Eq.~(\ref{final_elbo}). Thus, we can obtain the Monte Carlo estimate of the expectation of the GP auto-encoder from
\begin{align}\label{expect}
\hspace{-2mm}
\small
\mathbb{E}_{q(\vect{X}|\vect{Y})}
\left[ \log p(\vect{Y}^{(v)}|\vect{X})\right]\! = \!
\sum\nolimits_i \mathbb{E}_{\mathcal{N}(\vect{\xi}|\vect{0},\vect{I})}
\left[ \log p(\vect{y}_i^{(v)}|\hat{\vect{m}}_i + (\vect{S}_i^{1/2} + \hat{\sigma}_i\vect{I})\vect{\xi})\right].
\end{align}
The expectation of the ordinal classifier is computed in a similar manner.
The advantage of Eq.~(\ref{expect}) is twofold: (i) It allows for an efficient computation of the lower bound even when using arbitrary kernel functions (in contrast to~\cite{damianou2012manifold}); (ii) It provides an efficient, low-variance  estimator of the gradient~\cite{kingma2013auto}.
The extra approximation (via the expectation) in the gradient step requires stochastic gradient descent. We use AdaDelta~\cite{zeiler2012adadelta} for this purpose.

% \begin{algorithm}[t]
% \small
% \caption{MC-LVM: Learning and Inference}
% \label{alg}
% \begin{algorithmic}
% \\\hrulefill\State \textbf{Learning}
% \State Inputs: $\mathcal{D}=(\vect{Y}^{(v)}, \vect{Z}), v = 1, \ldots, V$
% \State Initialize $\vect{X}$ using PCA.%$\mu_{max}>>\mu_0>0$, $\rho=const.$, $\mathbf{X}_0$, $\mathbf{A}_0^{(v)}$, $\mathbf{\Lambda}_0^{(v)}$.
% \Repeat
% \State \textbf{Stage 1}

% Learn $\tilde{p}(\vect{x}) = p(\vect{x}|{{{\vect{y}}}^{(1)}},{\vect{y}^{(2)}})$ by training the specified GP.
% %$\theta_{X}^{(v)}$ by training the GP in Eq.~(\ref{GP_map})

% Draw $S$ latent variables $\vect{x}_s$ from $\tilde{p}(\vect{x})$

% \State \textbf{Stage 2}

% \textbf{E-step}: Use the current estimate of the parameters $\vect{\Theta}^{(old)}$ \\\qquad to compute the membership probabilities in Eq.~(\ref{bayes}).

% \textbf{M-step}: Update $\vect{\Theta}$ by maximizing Eq.~(\ref{eq4}).

% \State \textbf{Stage 3}

% Update the latent space using Eq.~(\ref{latent})
% \Until{convergence of Eq.~(\ref{eq4})}
% \State Outputs: $\vect{X}$, $\vect{\Theta}$
% \\\hrulefill
% \State \textbf{Inference}
% \State Inputs: $\vect{y}^{(1)}_\ast, \vect{y}^{(2)}_\ast$
% \State \textbf{Step 1:} Find the projection $\vect{x}_\ast$ to the latent space using Eq.~(\ref{gp_mu}).
% \State \textbf{Step 2:} Apply the logistic functions from Eq.~(\ref{sigmoid}) to the obtained embedding to compute the outputs $\vect{z}_\ast$.
% \State Output: $\vect{z}_\ast$
% \end{algorithmic}
% \end{algorithm}

Inference in the proposed method is straightforward: The test data $\vect{y}^{(v)}_\ast$, are first projected onto the manifold using the trained GP-encoder. In the second step, we apply the ordinal classifier to the obtained latent position.

\subsection{Relation to Prior Work on Gaussian Processes}
Our auto-encoder approach is inspired by neural-network counterparts proposed in~\cite{kingma2013auto,rezende2014stochastic}, where probabilistic distributions are defined for the input and output mapping functions. In the GP literature, auto-encoders are closely related to the notion of `back-constraints'. Back-constraints were introduced in~\cite{LawrenceC06} as a deterministic, parametric  mapping (commonly a multi-layer perceptron (MLP)) that pairs the latent variables of the GPLVM~\cite{lawrence2005probabilistic} with the observations. This mapping facilitates a fast inference mechanism and enforces structure preservation in the manifold. The same mechanism has been used to constrain the shared GPLVM~\cite{shon2006learning}, from one view in~\cite{ek2008gaussian} and multiple views in~\cite{eleftheriadis2015discriminative}.\looseness-1
%, in order to learn powerful generative models
%By including the back-constraints in the GPLVM formulation, the latent positions are indirectly obtained as a function of the observations, and thus, during training it is sufficient to optimize the parameters of the deterministic mapping. This has been proven advantageous for the ML optimization, since the number of parameters is significantly reduced.

%Similar line of works attempts have been introduced lately to constrain the GPLVM in a fully Bayesian training of the space.~\cite{damianou2015semi} proposed to train the B-GPLVM~\cite{titsias2010bayesian}, where the introduced variational distribution of the latent space was conditioned on some unobserved (or partially observed) inputs. 
Back-constraints have been recently introduced to the B-GPLVM~\cite{titsias2010bayesian}. In~\cite{damianou2015semi} the authors proposed to approximate the true posterior of the latent space by introducing a variational distribution conditioned on some unobserved  inputs. However, those inputs are not related to the observation space considered in this paper (\ie, the outputs $\vect{Y}$ of the GPLVM). In~\cite{dai2015variational} the variational posterior of the latent space is constrained by using the trick of the parametric deterministic mapping from~\cite{LawrenceC06}. Finally, in~\cite{eleftheriadis2015multi}, the authors replaced the variational approximation with a Monte Carlo expectation-maximization algorithm. Samples were obtained from the GP mapping from the observed inputs to the manifold.\looseness-1
%n attempt closer to ours has been proposed in~\cite{eleftheriadis2015multi}, where the authors replaced

%contrary to~\cite{LawrenceC06,ek2008gaussian,eleftheriadis2015discriminative,
%\dai2015variational},
Our proposed VGP-AE advances the current literature in many aspects: (1) We introduce a GP mapping for our recognition model. Hence, can model different uncertainty levels per input, which allows us to learn more confident latent representations. (2) The use of the non-parametric GPs also allows us to model complex structures at a lesser expense than the MLP (fewer parameters). Thus, it is less prone to overfitting and scales better to high-dimensional data. (3) Compared to~\cite{damianou2015semi} our probabilistic recognition model facilitates a low-dimensional projection of our observed features, while the variational constraint in~\cite{damianou2015semi} does not constitute a probabilistic mapping. (4) We learn the GP encoders/decoders in a joint optimization, while~\cite{eleftheriadis2015multi} train the two models in an alternating scheme.\looseness-1
%(GPs are regarded to be a generalization of an infinitely wide MLP~\cite{})
% borrowed the ideas from kingma/shakir
% sparse gp which if inducing are correctly selected can be very powerful/ expressive
% here we focus on one layer
% resubstitution allows for modeling complex kernel covariances
% Titsias used the backpropagation to learn hyperparmeters
%and is not a direct specific parameterization  in contrast to our well-defined GP model

\section{Experiments}
\vspace{-2mm}
We empirically assess the structure learning abilities of the proposed VGP-AE  as well as its efficacy when dealing with data of ordinal nature.
\vspace{-3mm}
\subsection{Experimental Protocol}
\paragraph{Datasets.} We first show the qualitative evaluation of the proposed VGP-AE on the MNIST~\cite{lecun1998mnist} benchmark dataset of images of handwritten digits. We use it to assess the properties of the auto-endoced manifold. We then show the performance of VGP-AE on two benchmark datasets of facial affect: DISFA~\cite{mavadati2013disfa}, and  BP4D~\cite{zhang2014bp4d} (using the publicly available data subset from the FERA2015~\cite{valstar2015fera} challenge). Specifically, DISFA contains video recordings of 27 subjects while watching YouTube videos. Each frame is coded in terms of the intensity of $12$ AUs, on a six-point ordinal scale. The FERA2015 database includes video of 41 participants. There are 21 subjects in the training and 20 subjects in the development partition. The dataset contains intensity annotations for $5$ AUs.
%Binghamton-Pittsburgh 4D Spontaneous Expression (ages 18-29)
%\ie, AU {\small(1, 2, 4, 5, 6, 9, 12, 15, 17, 20, 25, 26)}
%, \ie, AU {\small (6, 10, 12, 14, 17)}
%\vspace{-4mm}
\vspace{1mm}\\
\textit{Features.} In the experiment on MNIST dataset, we use the normalized raw pixel intensities as input, resulting in a 784D feature vector. For DISFA and FERA2015, we use both geometric and appearance features. Specifically, DISFA and FERA2015 datasets come with frame-by-frame annotations of 66 and 49 facial landmarks, respectively. After removing the contour landmarks from DISFA annotations, we end up with the same set of 49 facial points. We register the images to a reference face using an affine transform based on these points. We then extract Local Binary Patterns (LBP) histograms~\cite{ojala2002multiresolution} with 59 bins from patches centered  around each registered point. Hence, we obtain 98D (geometric) and 2891D (appearance) feature vectors, commonly used in modeling of facial affect.\looseness-1% We chose these features as they showed good performance in a variety of AU recognition tasks~\cite{senechal2012facial,eleftheriadis2015multi}. 
%\vspace{-4mm}
\vspace{1mm}\\
\textit{Evaluation.} As evaluation measures, we use the negative log-predictive density (NLPD) to assess the generative ability (reconstruction part) of our model. For the task of ordinal classification, we report the mean squared error (MSE) and the intra-class correlation (ICC(3,1))~\cite{shrout1979intraclass}. These are the standard measures for ordinal data. The MSE measures the classifier's consistency regarding the relative order of the classes. ICC is a measure of agreement between annotators (in our case, the ground truth of the AU intensity and the model's predictions). Finally, we adopt the subject-independent setting: for FERA2015 we report the results on the subjects of the development set, while for DISFA we perform a 9-fold (3 subjects per fold) cross-validation procedure.\looseness-1
%MSE is mainly used as a metric when dealing with continuous outputs.
%\vspace{-2mm}
\vspace{1mm}\\
\textit{Models.} We compare the proposed VGP-AE to the state of the art GP manifold learning methods that perform multi-input multi-output inference. These include: (i) manifold relevance determination (MRD)~\cite{damianou2012manifold}, a regression model based on variational inference, (ii) variational auto-encoded deep GP (VAE-DGP)~\cite{dai2015variational}, which uses a recognition model based on an MLP to constrain the learning of MRD, and (iii) multi-task latent GP (MT-LGP)~\cite{urtasun2008transferring}, which uses the same MLP-based recognition model and a maximum likelihood learning approach. We also compare to the variational GP for ordinal regression (vGPOR)~\cite{sheth2015sparse}. As a baseline, we use the standard GP~\cite{rasmussen2006gaussian} with a shared covariance function among the multi-outputs. We also compare to the single-output ordinal threshold model (SOR)~\cite{agresti2010analysis}. Finally, we compare to state of the art methods for joint estimation of AU intensity based on MRFs~\cite{sandbach2013markov} and latent trees (LT)~\cite{kaltwang2015latent}, respectively. For the single input (no fusion) methods  (GP, vGPOR, SOR, LT, MRF), we concatenate the two feature sets. The parameters of each method were tuned as described in the corresponding papers. For the GP subspace methods, we used the RBF kernel with ARD, and initialized with the 20D manifold. For the GP regression methods, we used the standard RBF. For the sparse variational GP methods (vGPOR, MRD, VAE-DGP) we used 200 inducing points, and 20 hidden units for the MLP in the recognition models of VAE-DGP and MT-LGP.\looseness-1

\subsection{Assessing the Recognition Model}
\begin{figure}[t]
\centering
\footnotesize
\setlength{\tabcolsep}{8pt}
\begin{tabular}{ccc} 
\includegraphics[scale=.17]{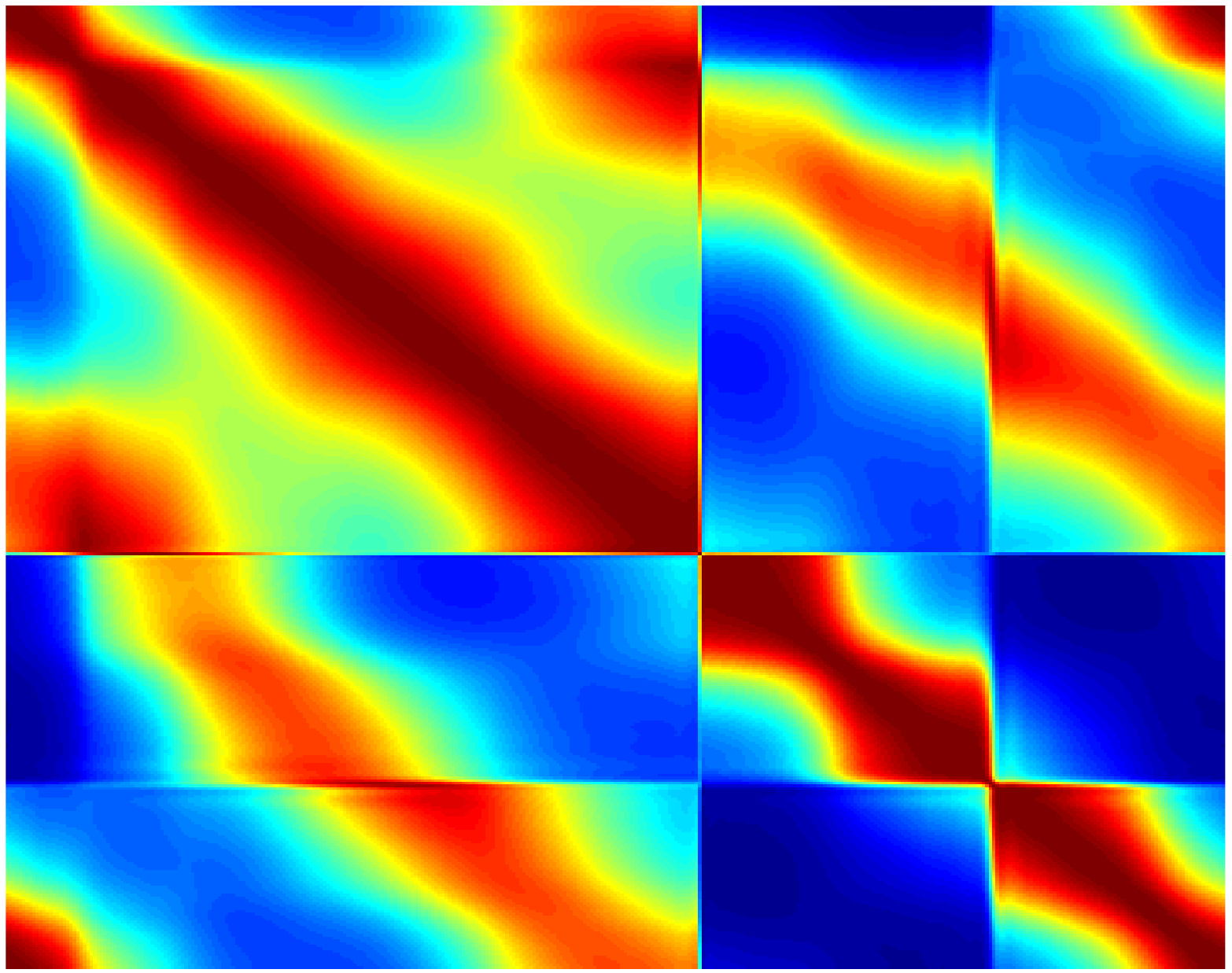}
&\includegraphics[scale=.17]{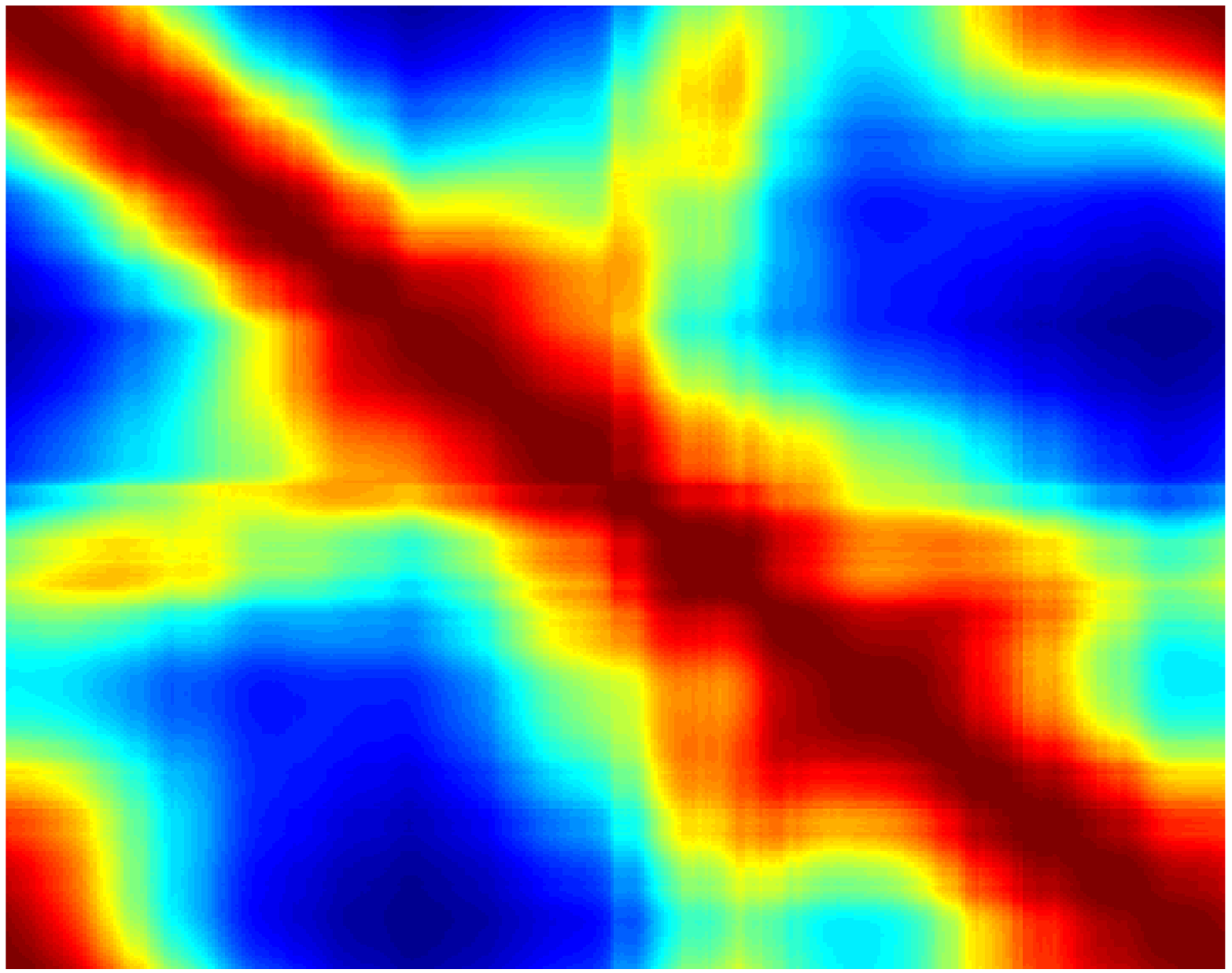}
&\includegraphics[scale=.17]{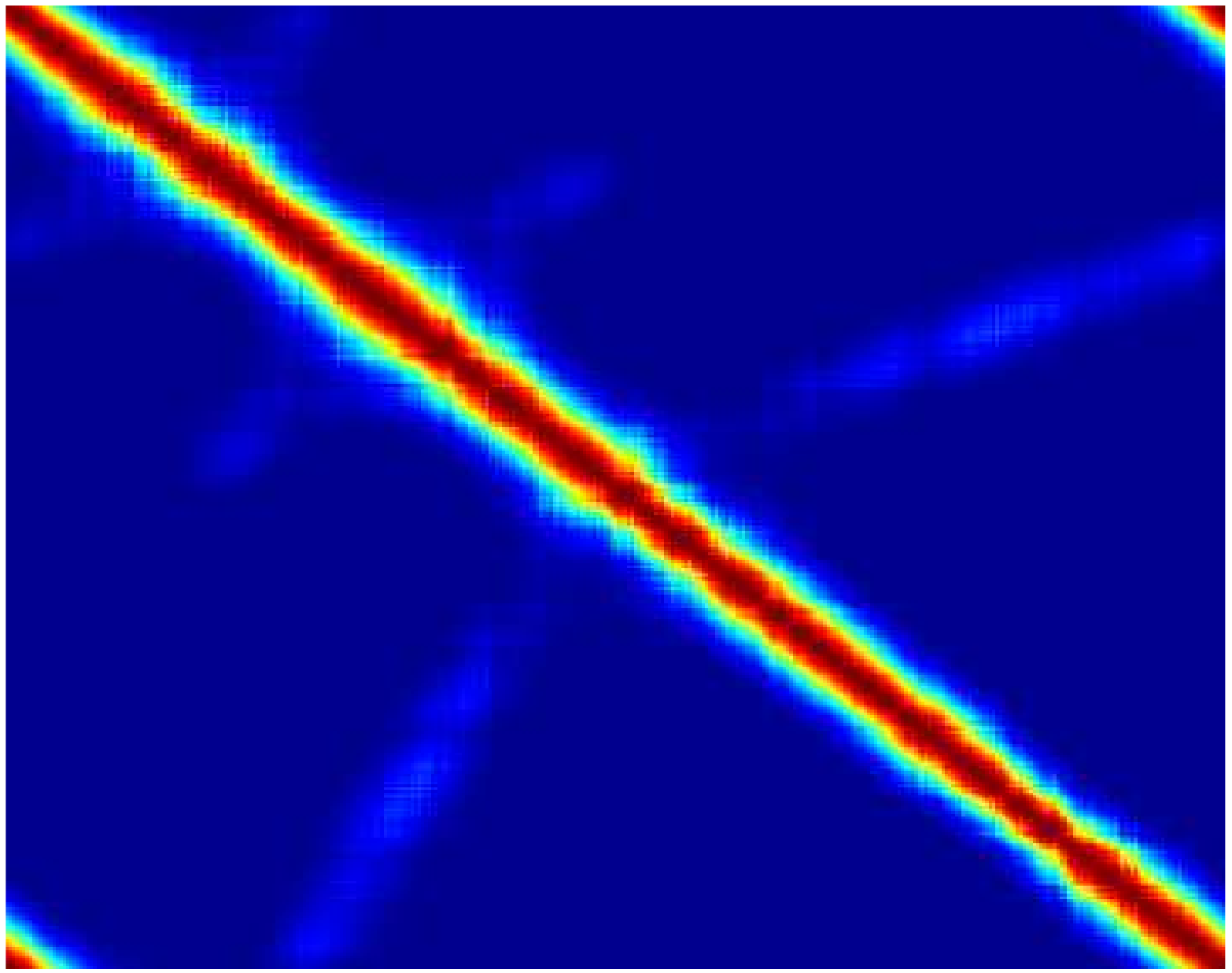}
\\
\includegraphics[scale=.25]{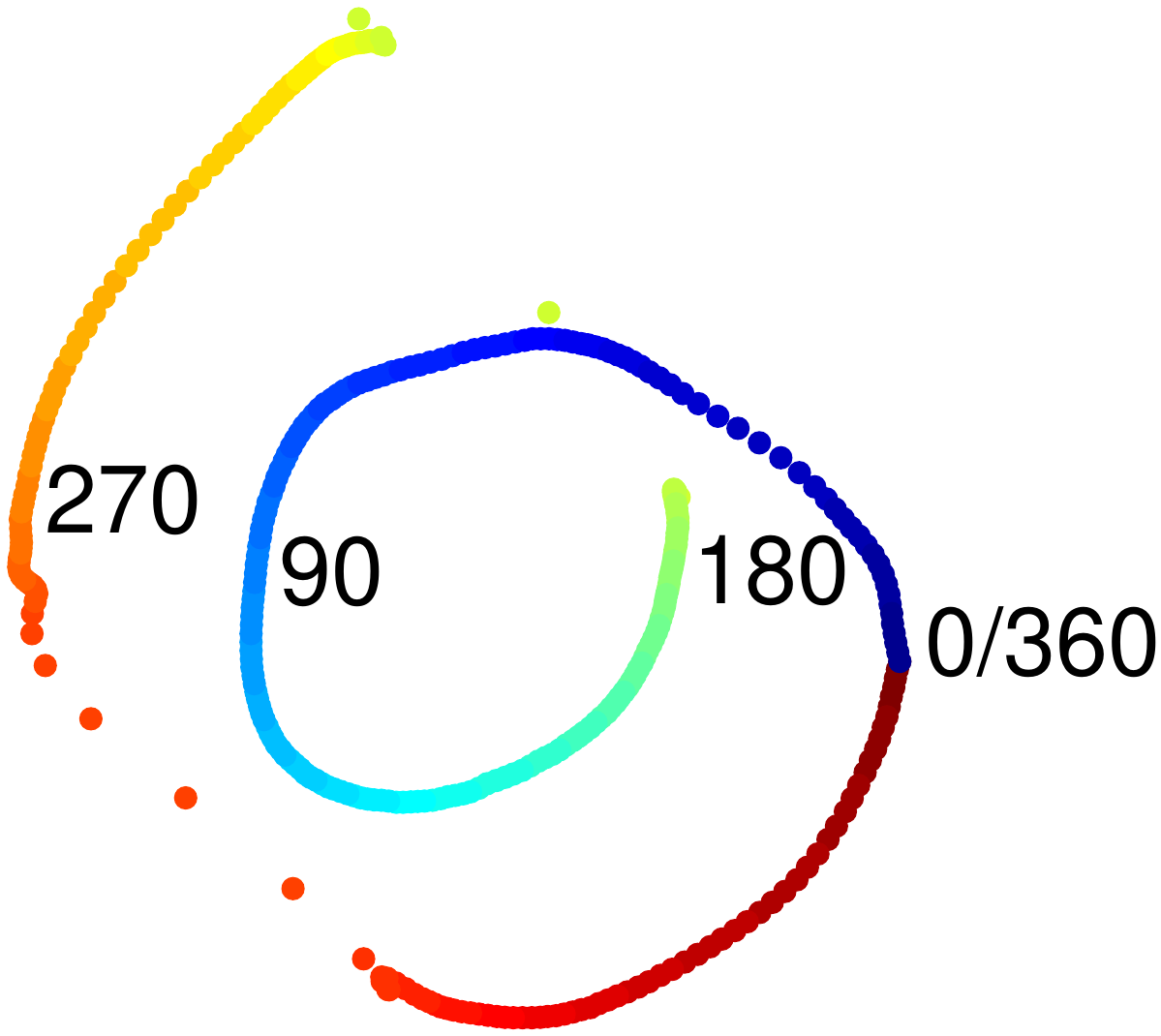}
&\includegraphics[scale=.25]{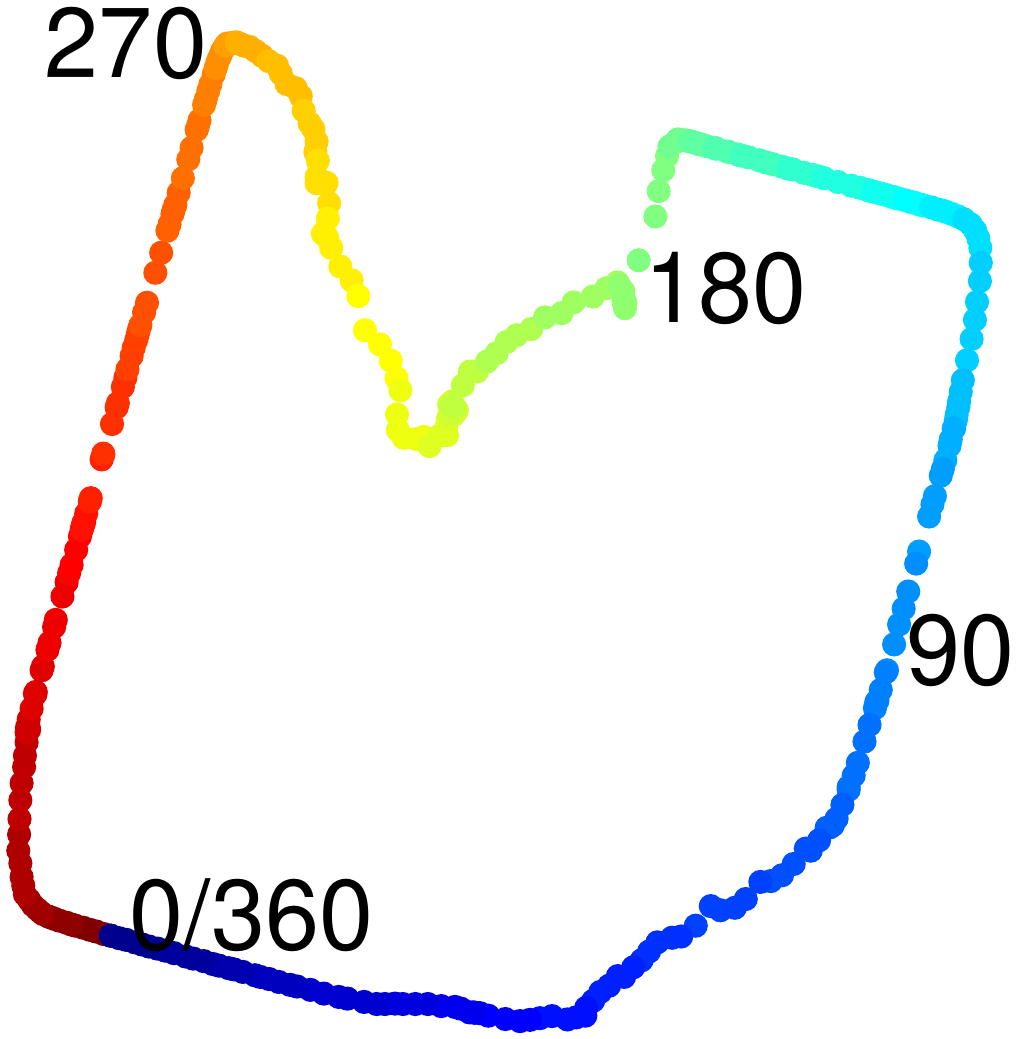}
&\includegraphics[scale=.28]{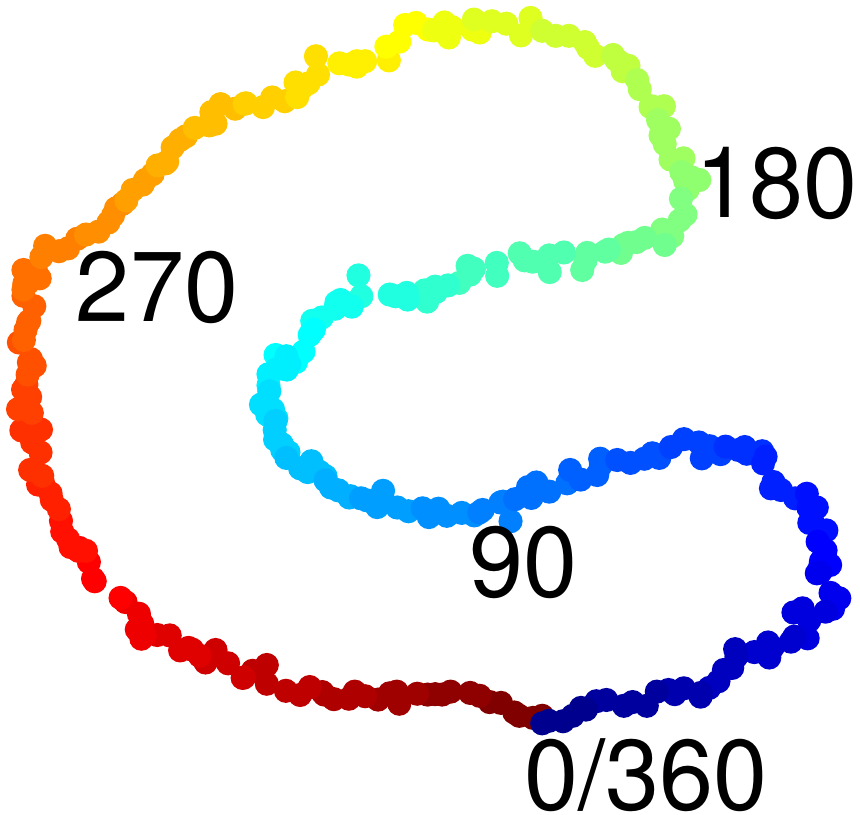}
\\
B-GPLVM~\cite{titsias2010bayesian} & VAE-DGP~\cite{dai2015variational} & VGP-AE
\end{tabular}
\caption{{\footnotesize Recovering the structure of a rotated `$1$' from MNIST. The learned kernel matrices (upper row) and 2D manifolds (lower row) obtained from B-GPLVM (left), VAE-DGP (middle) and the proposed VGP-AE (right), initialized from the same random instance.}} %The ellipses around the latent coordinates account for 2 times the s.d. of the variational distribution $q(\vect{X})$.}}
\label{fig_toy}
\end{figure}
In the following, we qualitatively assess the benefits of the proposed recognition model in the task of manifold recovery from the MNIST dataset. We select an image depicting the digit `1' and rotate it around $360^\circ$. This results in a set of images of `1's rotated at a step of $1^\circ$. Our goal is to infer the true structure of the data, for which we know {\em a priori} that it should correspond to a diagonal-like kernel and a circular manifold. However, the challenge arises from the symmetry of digit `1', which is almost identical at opposite degrees (\eg, $0^\circ$ and $180^\circ$). The results are depicted in Fig.~\ref{fig_toy}. Note that since we do not deal with the classification task we exclude the ordinal component in VGP-AE. We compare the learned manifold structure to the B-GPLVM~\cite{titsias2010bayesian}, which does not model the back-projection to the latent space, and a single layer VAE-DGP, where the back-projections are modeled using MLP. In Fig.~\ref{fig_toy}(upper row), we see from the learned kernels that the B-GPLVM is unable to fully unravel the dissimilarity between the `inverted' images, resulting also in a non-smooth kernel with a discontinuity at $180^\circ$ and $270^\circ$. By contrast, the VAE-DGP benefits from the recognition model and manages to resolve this to some extent. Yet, the recovered kernel still suffers from a discontinuity around $180^\circ$. On the other hand, the proposed VGP-AE, by using the more general recognition model based on GPs (infinitely wide MLP), succeeds to accurately discover the true underlying manifold, also resulting in a more smooth, almost ideal kernel. These observations are further supported by the instances of the learned 2D manifolds in Fig.~\ref{fig_toy}(lower row). B-GPLVM learns a disconnected manifold with `jumps' at $180^\circ$ and $270^\circ$. However, both the VAE-DGP and proposed VGP-AE recover a circular manifold, with the manifold recovered by VGP-AE being more symmetric. 
%On the contrary, the proposed method recovers a spiral of two homocentric circles, where the latent coordinates that correspond to opposing degrees are equidistant on the manifold. Note also the difference in the estimated variances (circles around the latent points) between B-GPLVM and the proposed. The latter learns confident projections for the points on the circles, and the variance increases at the crossing point. B-GPLVM falsely estimates a high variance for all the latent positions.
% This linear mapping on the inputs, results on a highly non-linear transformation

% \begin{figure}[t]
% \centering
% \footnotesize
% \setlength{\tabcolsep}{5pt}
% \begin{tabular}{cccc} 
% \multicolumn{2}{c}{(a) Learned kernels} & \multicolumn{2}{c}{(b) Learned manifolds}\\
% \includegraphics[scale=.16]{kernel_mrd_toy-cropped.pdf}
% &\includegraphics[scale=.16]{kernel_toy-cropped.pdf}
% &\includegraphics[scale=.18]{latent_mrd_toy-cropped.pdf}
% &\includegraphics[scale=.18]{latent_toy-cropped.pdf}\\
% B-GPLVM~\cite{titsias2010bayesian} & ours & B-GPLVM~\cite{titsias2010bayesian} & ours
% \end{tabular}
% \caption{{\footnotesize Recovering the structure of rotated `$1$' from MNIST dataset. The learned kernels (a) and manifolds (b) obtained from B-GPLVM~\cite{titsias2010bayesian} (left) and the proposed method (right), initialized from the same 2-D latent space. The ellipses around the latent coordinates account for 2 times the s.d. of the variational distribution $q(\vect{X})$.}}
% \label{fig_toy}
% \end{figure}

\subsection{Convergence Analysis}
\begin{figure}[t]
\centering
\footnotesize
\setlength{\tabcolsep}{3pt}
\begin{tabular}{ccc}
\includegraphics[scale=.24]{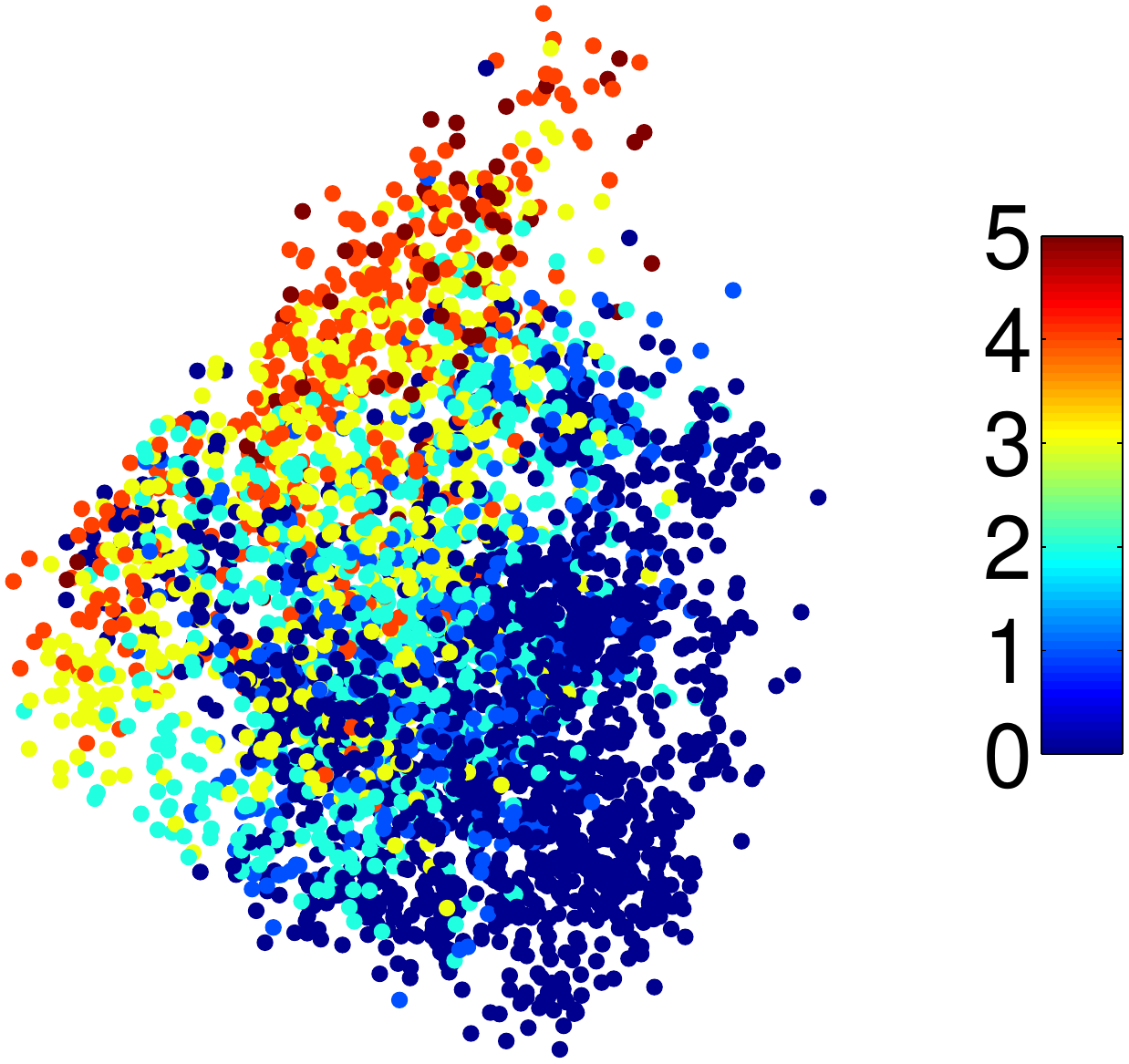}
 & \includegraphics[scale=.20]{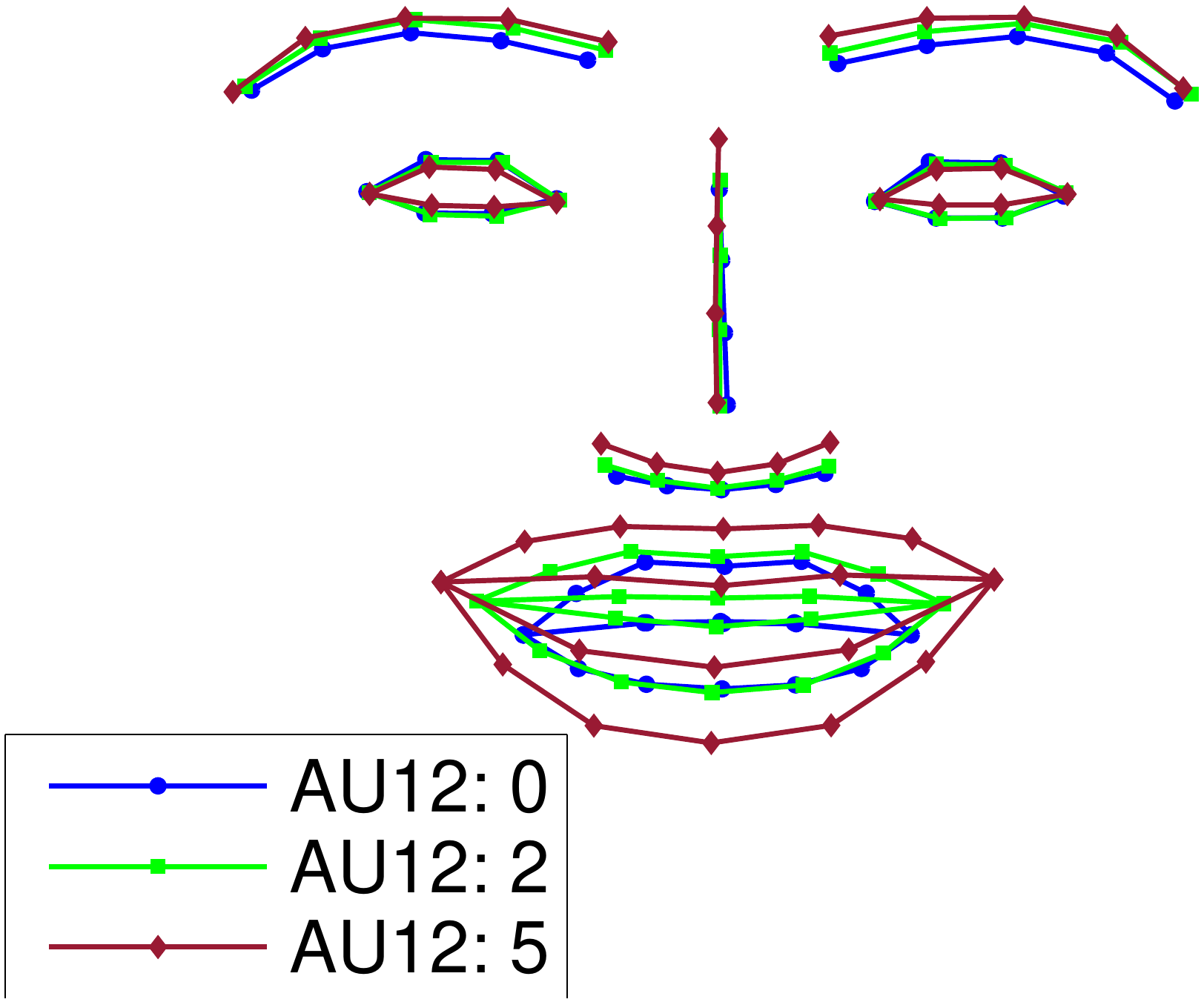}
&\includegraphics[scale=.20]{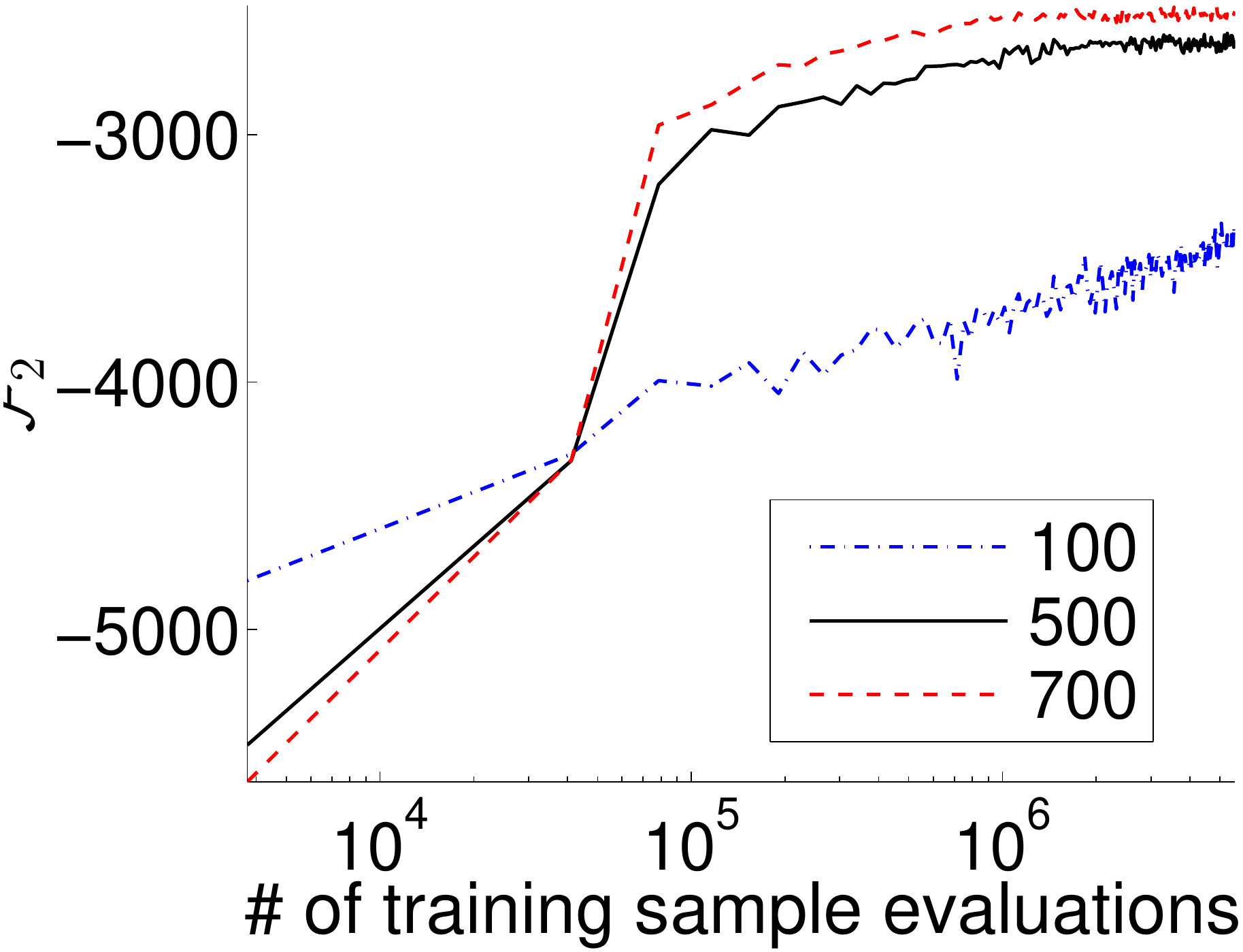}\\%_new
(a) Latent space & (b) Generated faces & (c) Lower bound\\
\includegraphics[scale=.20]{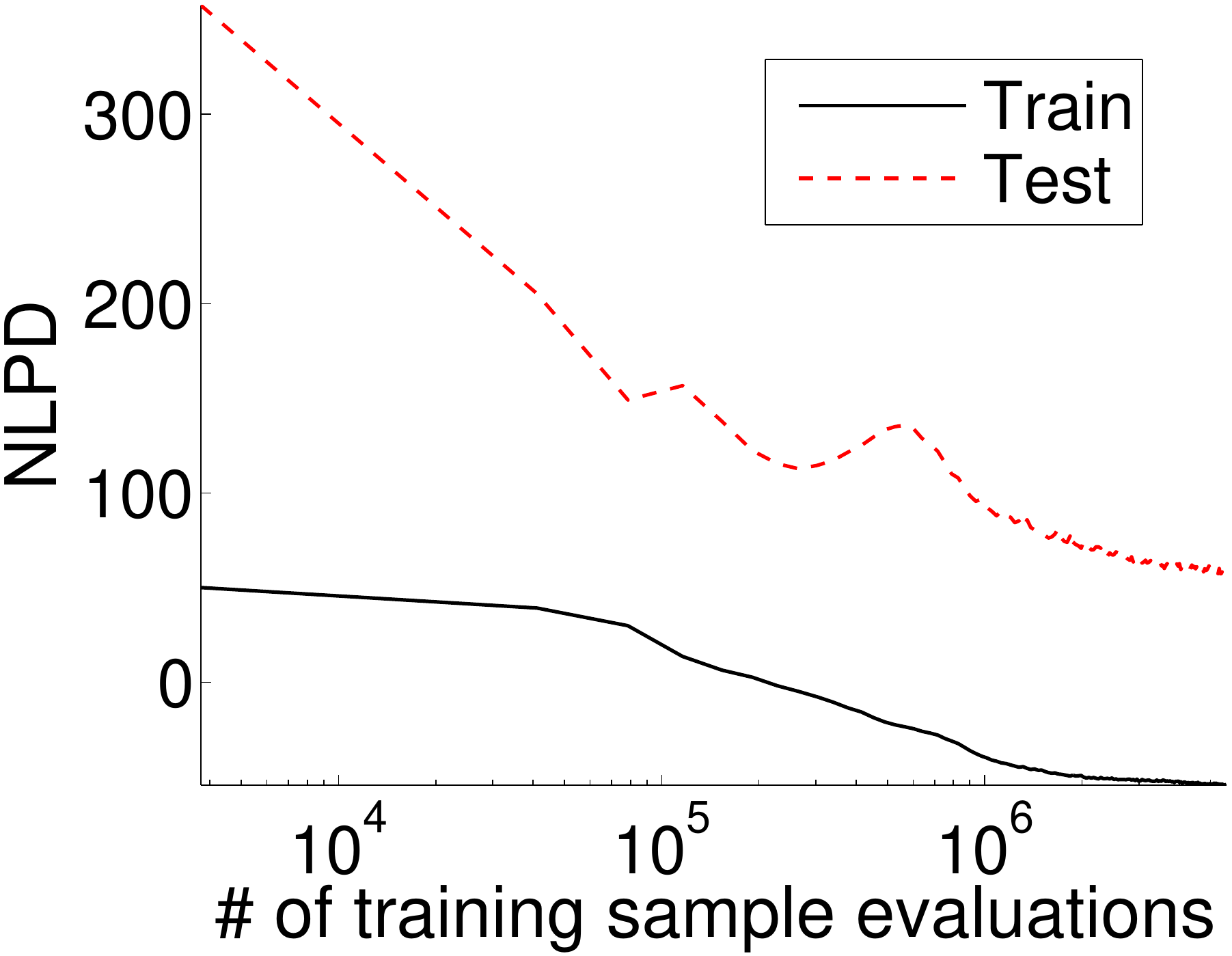} &\includegraphics[scale=.20]{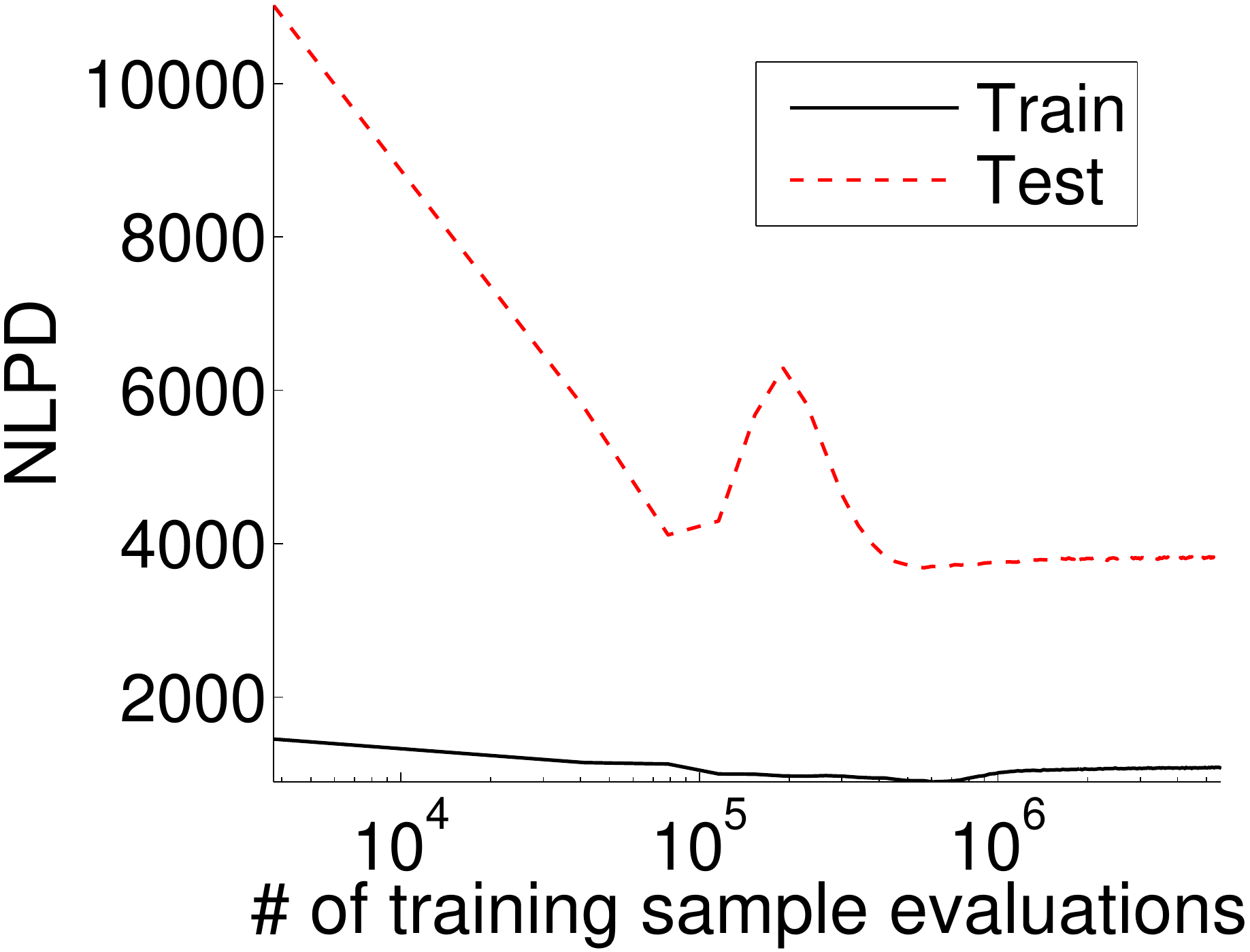}
&\includegraphics[scale=.20]{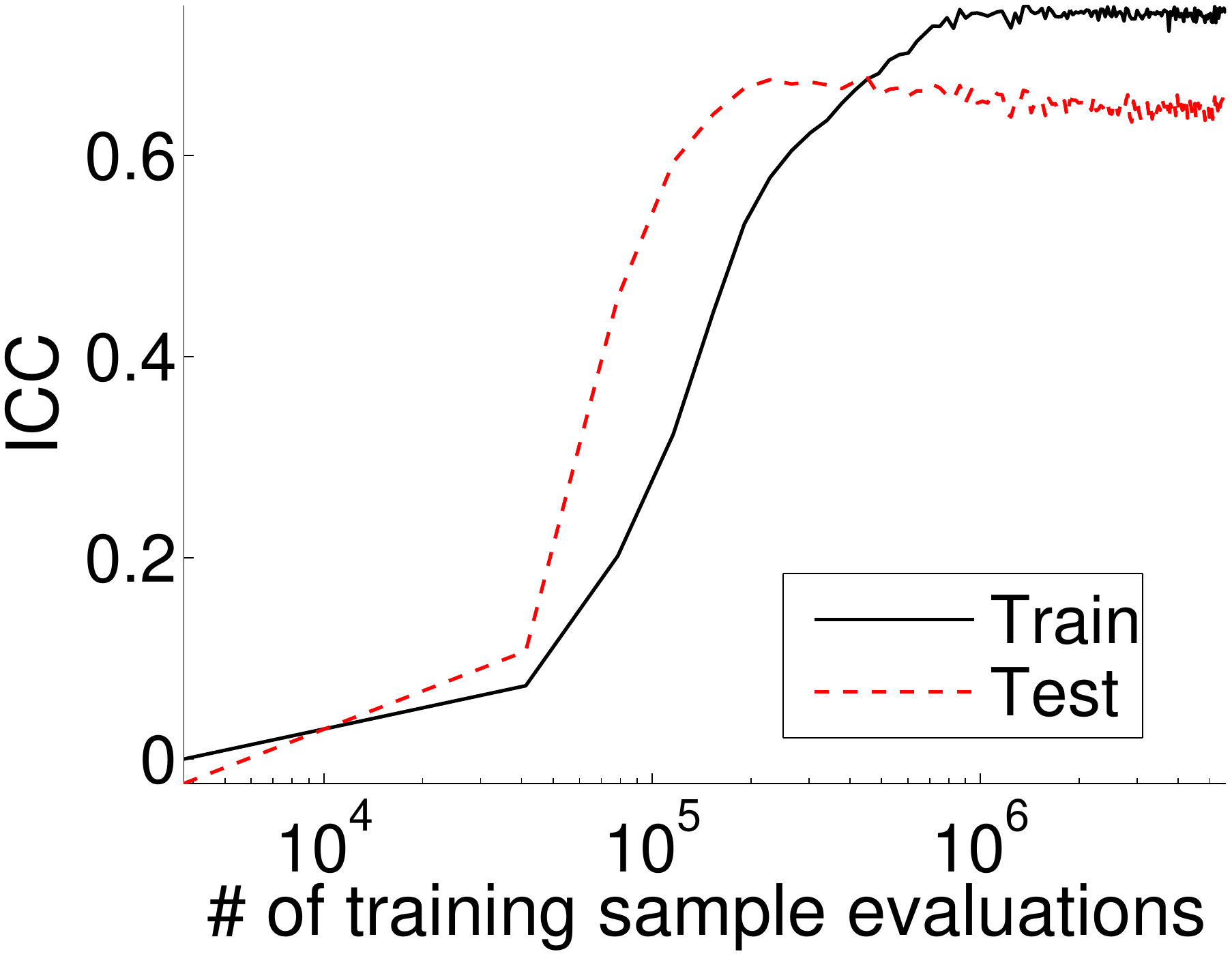}\\
(d) Reconstruction of points & (e) Reconstruction of LBP & (f) Predictions
\end{tabular}
\caption{{\footnotesize Convergence analysis of the proposed method on FERA2015. 
(a) The recovered latent space with ordinal information from AU$12$, and (b) reconstructed face shapes sampled from different regions of the manifold. (c) the estimated average variational lower bound, $\mathcal{F}_2$, per datapoint, for different batch sizes. The model's reconstruction capacity for the points (d) and LBP (e) features, measured by the NLPD. (f) the average ICC for the joint AU intensity estimation. The horizontal axis corresponds to the amount of training points evaluated after $1500$ epochs of the stochastic optimization.}}
\label{fig_conv}
\end{figure}
We next demonstrate the convergence of VGP-AE in the task of AU intensity estimation on FERA2015. Fig.~\ref{fig_conv}(a) shows the effect of learning the ordinal classifier and the auto-encoded manifold within the joint optimization framework. It can be clearly seen from the recovered space that the information from the labels has been correctly encoded in the manifold, which now has an ordinal structure (the depicted coloring accounts for the `ordinality' of AU12). As depicted in Fig.~\ref{fig_conv}(b), we can accurately reconstruct face shapes with different AU intensities, by sampling from different regions of the space. 
Fig.~\ref{fig_conv}(c) shows the convergence of the proposed method when optimizing the lower bound $\mathcal{F}_2$ of Eq.~(\ref{final_elbo}) for different batch sizes of the stochastic optimization. With a small batch size (100 datapoints) the model cannot estimate  the structure of the inputs well. Hence, it approximates the log-marginal likelihood less accurately. By increasing the batch size to 500, the model converges to a better solution and optimization becomes more stable since the curve becomes smoother over the iterations. Further increase of the batch size does not have a considerable effect.% on the training process. 

In Fig.~\ref{fig_conv}(d)--(e) we evaluate the generative part of the auto-encoder by measuring the model's ability to reconstruct both input features (points and LBPs) in terms of NLPD. First of all, it is clear that our Bayesian training prevents the model from overfitting, since the NLPD of the test data follows the trend of the training data. Furthermore, we can see that the model can reconstruct the geometric features  better than the appearance, which is evidenced by the lower NLPD (around $-50$ for points and $1500$ for LBPs). We partly attribute this to the fact that the LBPs are of higher dimension and therefore more difficult to reconstruct. Another reason for this difference is that the model learns to reconstruct the part of the features that enclose the more relevant information regarding the task of classification. The latter is further supported by Fig.~\ref{fig_conv}(e), where we see the progress of the average ICC during the optimization. In the beginning, the model has no information since the latent space is initialized randomly. As we progress  the model fuses the information of the input features in the latent space and unravels the structure of the data. Thus, ICC starts rising and reaches its highest value, $.65$ on the test data. After that point the model does no longer benefit from the appearance features: it has reached the plateau. %Finally, another interesting observation from Fig.~\ref{fig_conv}(e) is the higher ICC on the test data at the beginning of the optimization. This is partly attributed to the auto-encoding process. The ICC for the train data is evaluated on the learned latent positions, while the one for the test data is after using the recognition model to project to the latent space. Hence, this mapping serves also as an extra de-noising step. 

% \multirow{2}{*}{
% \smash{\raisebox{-31pt}{\includegraphics[scale=.22]{latent-cropped.pdf}}}}
% \smash{\raisebox{31pt}{(a)}}

\subsection{Model Comparisons on Spontaneous Data of Facial Expressions}
 
\begin{table}[t]
\setlength{\tabcolsep}{1.1pt}  
\center
\caption{Joint AU intensity estimation on DISFA and FERA2015.}
\scalebox{0.85}{
\begin{tabular}{l|l|cccccccccccc|c|ccccc|r}%{1\linewidth}
    \hline
    \rowcolor{Gray}
        \hline
\multicolumn{2}{l|}{\quad Dataset} & \multicolumn{13}{c|}{{DISFA}} & \multicolumn{6}{c}{{FERA2015}}\\
	\rowcolor{Gray}
    \multicolumn{2}{l|}{\quad AU}  & 1            & 2            & 4            & 5            & 6            & 9            & 12           & 15           & 17           & 20           & 25           & 26           & Avg.         & 6            & 10            & 12            & 14            & 17			& Avg.         \\
    \hline
    \toprule
    \parbox[t]{2mm}{\multirow{11}{*}{\rotatebox[origin=c]{90}{\qquad ICC}}}
    & VGP-AE  & .48          & .47 & .62          & .19 & \textbf{.50}          & \textbf{.42}          & \textbf{.80}           & .19 & \textbf{.36}          & .15          & \textbf{.84}          & .53 & \textbf{.46} & \textbf{.75}  & \textbf{.66} & \textbf{.88} & .47		& \textbf{.49}   & \textbf{.65}       \\
    & VAE-DGP~\cite{dai2015variational}  & .39          & .34 & .46          & .13 & .40          & .31          & .75           & .14 & .23          & .14          & .75          & .45 & .38 & .72  & .61 & .82 & .40		& .38   & .59       \\
    & MRD~\cite{damianou2012manifold}  & .46          & .39 & .43          & .09 & .28          & .34          & .71           & .09 & .30          & .09          & .73          & .36 & .36 & .68  & .59 & .80 & .38		& .38   & .57       \\
    & MT-LGP~\cite{urtasun2008transferring}  & .41          & .33 & .28          & .10 & .23          & .22          & .56           & .13 & .26          & .18          & .65          & .23 & .30 & .67  & .61 & .80 & .37		& .41   & .57       \\
    & vGPOR~\cite{sheth2015sparse}     & \textbf{.53}          & \textbf{.49}           & .54 & .21           & .35 & .40          & .75 & .18          & .30          & .16          & .79          & .39          & .42          & .74          & .62          & .84          & \textbf{.48}           & .35          & .61\\
    & GP~\cite{rasmussen2006gaussian}     & .28          & .13           & .42 & .03           & .13 & .23          & .62 & .08          & .26          & .19           & .67          & .23          & .27          & .69          & .58          & .81          & .35           & .38          & .56\\
    & SOR~\cite{agresti2010analysis}    & .25          & .18          & \textbf{.65}          & .08          & .46          & .15          & .77          & .14          & .24          & .04           & .82 & \textbf{.57}          & .36          & .61          & .50          & .77          & .28          & .45          & .52\\
%     & SVR~\cite{smola1997support}    & .33          & .25          & .37          & .26          & .30          & .22          & .52          & .23          & .27           & .15          & .62          & .37          & .32          & .46          & .41          & .56          & .26          & .31          & .40\\
    & LT~\cite{kaltwang2015latent}     & .28          & .26          & .44          & .24          & .50          & .13          & .69          & .06          & .21           & .06          & .62          & .37          & .32          & .70          & .59          & .76          & .30          & .31          & .53\\
    & MRF~\cite{sandbach2013markov}    & .46          & .38          & .50           & \textbf{.37}          & .41          & .34           & .67          & \textbf{.32}          & .29          & \textbf{.20}          & .69          & .46          & .42          & .64          & .53          & .79          & .34          & .46           & .55\\
    \hline
    \hline
    \parbox[t]{1mm}{\multirow{11}{*}{\rotatebox[origin=c]{90}{\qquad MSE}}}    
    & VGP-AE  & .51          & \textbf{.32} & 1.13          & .08 & .56          & .31 & .47 & .20 & \textbf{.28}          & .16 & \textbf{.49} & \textbf{.44} & .41 & \textbf{.82} & \textbf{1.28} & \textbf{.70} & \textbf{1.43} & \textbf{.77} & \textbf{1.00}\\
    & VAE-DGP~\cite{dai2015variational}  & .40          & .36 & .95          & .08 & .48          & .29 & \textbf{.43} & .19 & .32          & .16 & .76 & \textbf{.44} & .41 & .91 & 1.33 & .81 & 1.46 & .86 & 1.07\\    
   & MRD~\cite{damianou2012manifold}  & .42          & .38 & 1.31          & .08 & .56          & \textbf{.27} & .47 & .20 & .36          & .18 & .82 & .53 & .46 & 1.00 & 1.39 & .83 & 1.64 & .88 & 1.15\\
   & MT-LGP~\cite{urtasun2008transferring}  & .40          & .35 & 1.25          & .08 & .60          & .30 & .73 & .18 & .36          & .16 & 1.19 & .67 & .52 & .97 & 1.31 & .81 & 1.58 & .84 & 1.10\\
   & vGPOR~\cite{sheth2015sparse}     & .38          & .34          & .95 & \textbf{.06}          & .57 & \textbf{.27}          & \textbf{.43} & .18          & .33 & .18          & .65          & .53          & .41 & 1.00 & 1.54 & .76 & 1.78          & 1.11          & 1.24\\
   & GP~\cite{rasmussen2006gaussian}     & .52          & .51          & 1.13 & .13          & .65 & .36          & .61 & .23          & .38 & .20          & .94          & .66          & .53 & .94 & 1.40 & .76 & 1.62          & .88          & 1.12\\
    & SOR~\cite{agresti2010analysis}    & .47 & .40 & 1.13          & .07          & .63          & .37           & .55          & .21 & .35           & .21           & .71          & .61           & .48          & 1.44          & 1.82          & 1.08          & 2.58          & 1.01          & 1.59\\
%     & SVR~\cite{smola1997support}    & .42          & .38          & .95          & .07          & .45          & .31          & .56          & .15 & .30          & .16          & .81          & \textbf{.44}          & .42          & 1.19          & 1.44          & 1.26          & 1.60          & 0.83          & 1.27\\
    & LT~\cite{kaltwang2015latent}     & .44           & .38          & \textbf{.93}          & \textbf{.06} & \textbf{.36}          & .32          & .46          & .16          & .29          & \textbf{.15}          & .97          & \textbf{.44}          & .41          & .89          & 1.33          & .91          & 1.48          & .85          & 1.09\\
    & MRF~\cite{sandbach2013markov}    & \textbf{.37}          & .35          & .94           & \textbf{.06}          & .45          & .29          & .46          & \textbf{.13}          & .32          & .16          & .77          & \textbf{.44}          & \textbf{.40}           & 1.20          & 1.66          & .86           & 2.19          & .92          & 1.37\\
\hline
\hline
\end{tabular}
}
\label{tab}
\end{table} 
 
We compare the proposed approach to several methods on the spontaneous data from the DISFA and FERA2015 datasets. Table~\ref{tab} summarizes the results. First, we observe that all methods perform significantly better (in terms of ICC) on the data from FERA2015 than on DISFA. This is mainly due to the fact that FERA2015 contains a much more balanced set of AUs (in terms of activations), and hence, all models (single- and multi-output) can learn the classifiers for the target task better.  Furthermore, our proposed approach performs significantly better than the compared GP manifold learning methods, which treat the output labels as continuous variables. MRD lacks the modeling of back-projections. This results in learning a less smooth manifold of facial expressions, which affects its representation abilities, and hence, its predictions. On the other hand, the VAE-DGP learns explicitly the mapping from the observed features to the latent space in a deterministic and parametric fashion. Although this strategy is proven to be superior to unconstrained learning, it can be severely affected in cases where we have access to noisy and high-dimensional features. MT-LGP also models the back-mappings. However, it reports worse results, especially on DISFA. This drop in the performance is accounted to the non-Bayesian learning of the manifold, which constitutes the model more prone to overfitting.

%However, this is also the reason why all models report higher MSE on FERA2015. Since we have more AU activations on FERA2015, the cost of falsely predicting the ordinal state of each AU becomes higher, and thus, the MSE grows.

%Both MRD and MT-LGP treat the output labels as a continuous variable, a fact which eventually has a negative impact on the task of intensity estimation.
%Moreover, MT-LGP is trained on a non-Bayesian manner (it uses the MAP learning strategy), and thus, it is more likely to overfit the training data, a fact which is evidenced by the low ICC score on DISFA dataset.

Regarding the sparse ordinal regression instance of GPs, \ie, vGPOR, we  see that it manages to learn relatively accurate mappings between features and labels, and thus, performs close to our proposed method. However, it reports worse results since it cannot achieve the desirable fusion of the features without learning an intermediate latent space. The baseline methods, \ie, GP and SOR, report lower results. The GP attains low scores due to handling the ordinal outputs in a continuous manner while the ordinal modeling helps SOR to report consistently better. %However, the differences in the structure of the two modalities cannot be explained well from the parametric model of the SOR. This results in falsely predicting the scale of the output labels, which is evidenced by the high MSE. 

Finally, the proposed approach significantly outperforms the state of the art methods in the literature of AU intensity estimation, \ie, LT and MRF. LT learns the label information in a generative manner, and treats them as extra feature dimensions. Although this approach can be beneficial in the presence of noisy features~\cite{kaltwang2015latent}, it suffers from learning complicated and large tree structures when falsely detecting connections between features and AUs. Hence, it performs worse. The MRF performs on par to the proposed method on DISFA and achieves the best average MSE, but it is consistently worse on FERA2015. This inconsistency  is due to its two-step  learning strategy, which results in unraveling a graph that cannot explain simultaneously all different features and AUs.
%(first model the AUs and then learn the graph connections)
% vGPR also attains lower score as it also treats the outputs in a continuous manner. Similar findings can be observed by the results of the baseline methods, \ie, SOR and SVR. Again, the ordinal modeling helps SOR to report higher ICC scores compared to SVR. However, the differences in the structure of the two modalities cannot be explained well from the parametric model of the SOR. This results in falsely predicting the scale of the output labels, which is evidenced by the high MSE.

\begin{figure}[t]
\centering
\footnotesize
\includegraphics[scale=.68]{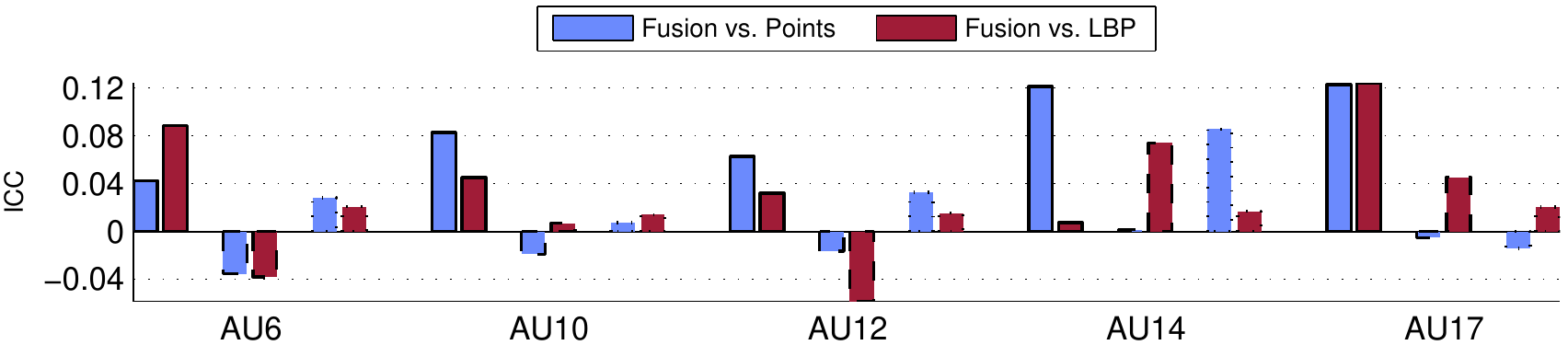}%bar_new
\caption{{\footnotesize Demonstration of the gain/loss from feature fusion for joint AU intensity estimation on FERA2015. Within each AU the first tuple (solid line) corresponds to the proposed VGP-AE, the second tuple (dashed line) to the VAE-DGP~\cite{dai2015variational}, and the third tuple (dotted line) to the vGPOR~\cite{sheth2015sparse}.}}
\label{fig_bar}
\end{figure}

In Fig.~\ref{fig_bar} we evaluate the attained fusion between the best performing methods on FERA2015, \ie, the proposed VGP-AE, VAE-DGP~\cite{dai2015variational} and vGPOR~\cite{sheth2015sparse}. As we can
see, the proposed approach (solid line, first tuple) manages to accurately fuse the information from the two input features in the learned manifold. Thus, it achieves higher ICC on all AUs compared to when the two modalities are used individually as input features. On the other hand, although vGPOR (third tuple, dotted line) reports also high ICC scores, it does not benefit from the presence of the two features: In most cases it cannot achieve a significant increase compared to the individual inputs. Finally, VAE-DGP (middle tuple, dashed line) consistently attains better performance on all AUs with a single feature as input. This can be attributed to modeling the recognition model via the parametric MLP. The latter affects the learning of the manifold, especially when dealing with the high-dimensional noisy appearance features.

\begin{figure}[tb]
\centering
\footnotesize
\setlength{\tabcolsep}{2pt}
\begin{tabular}{cccccc}
\rot{\rlap{\qquad VGP-AE}} & 
\includegraphics[scale=.17]{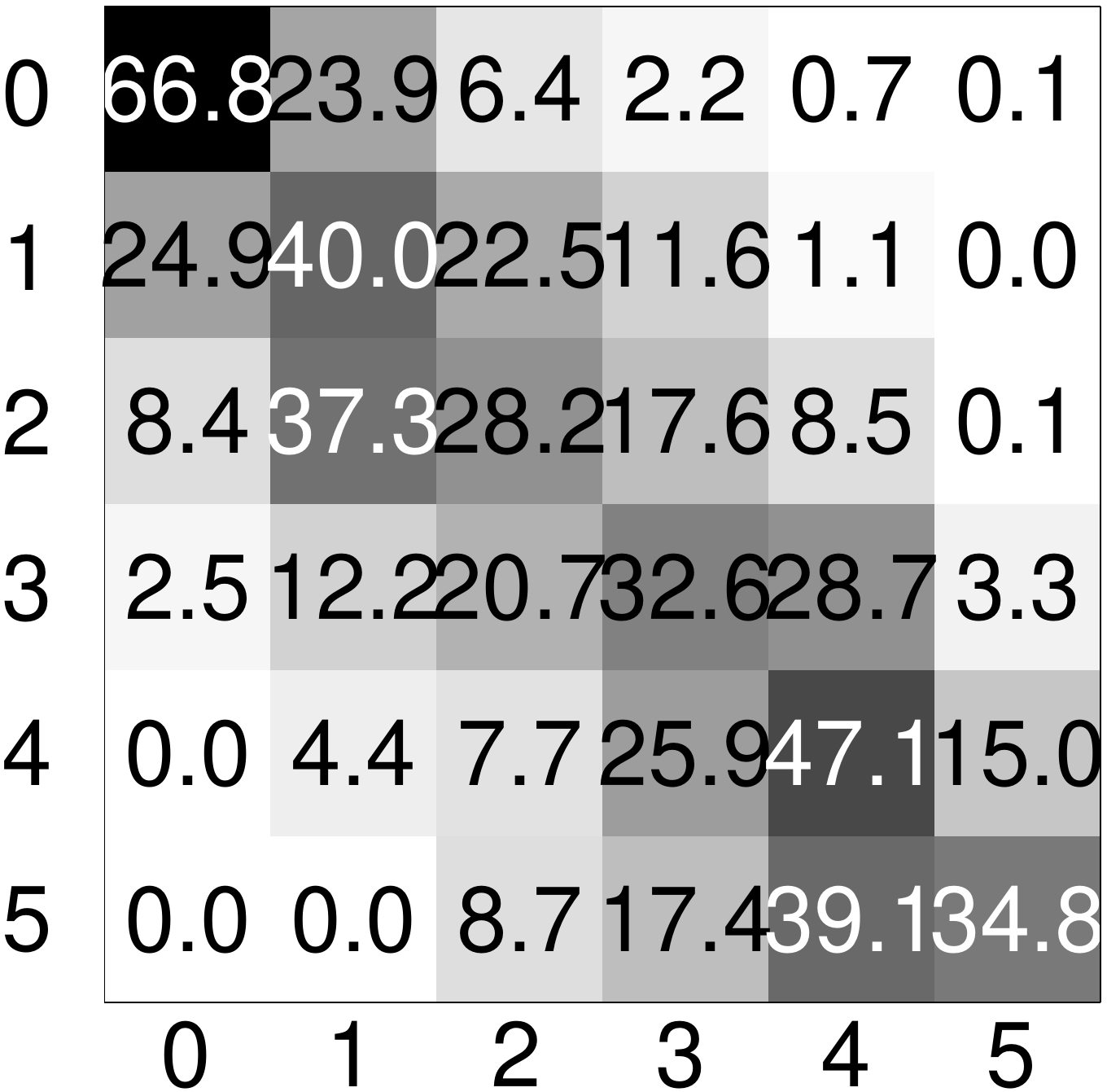} & \includegraphics[scale=.17]{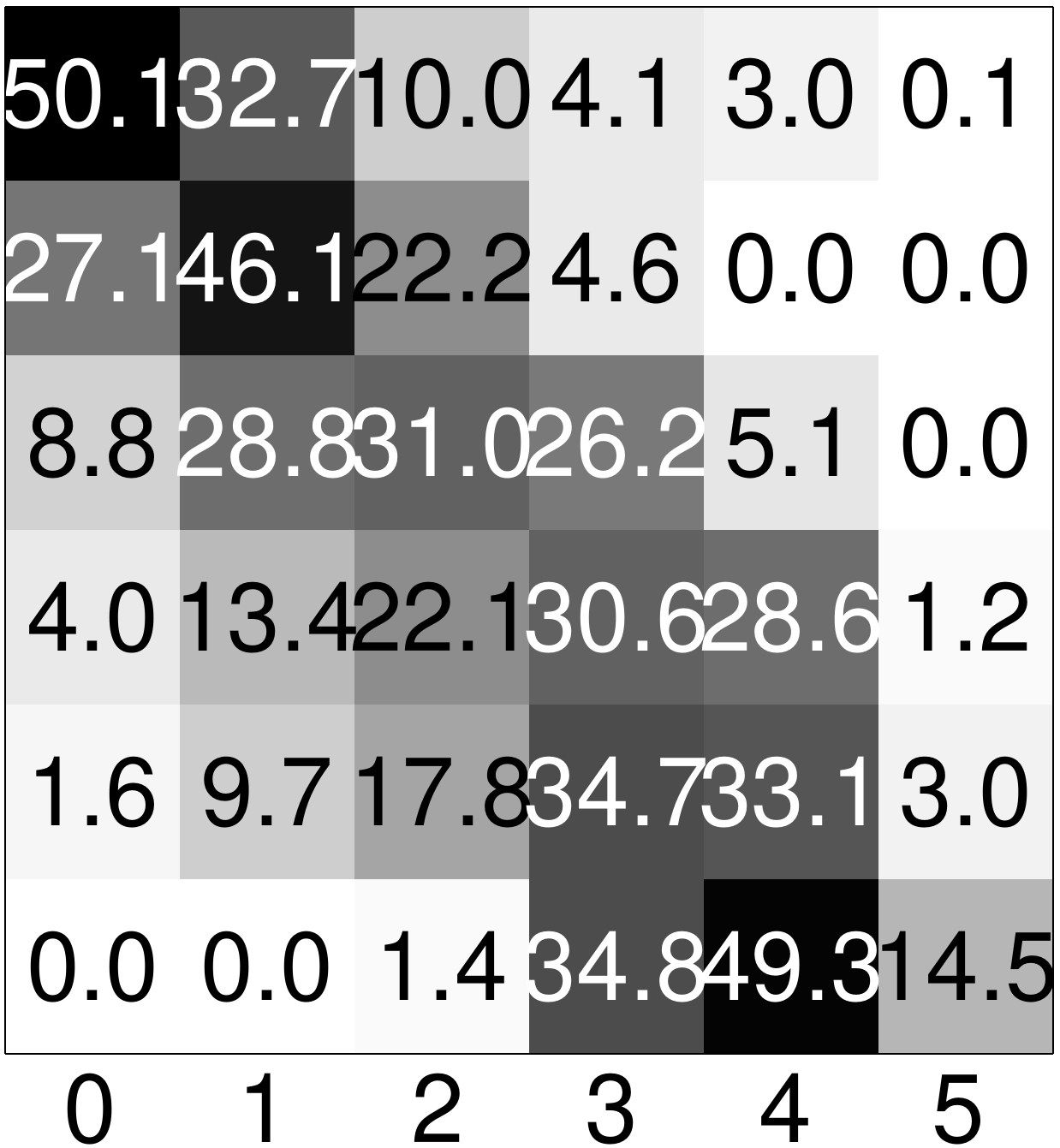}
&\includegraphics[scale=.17]{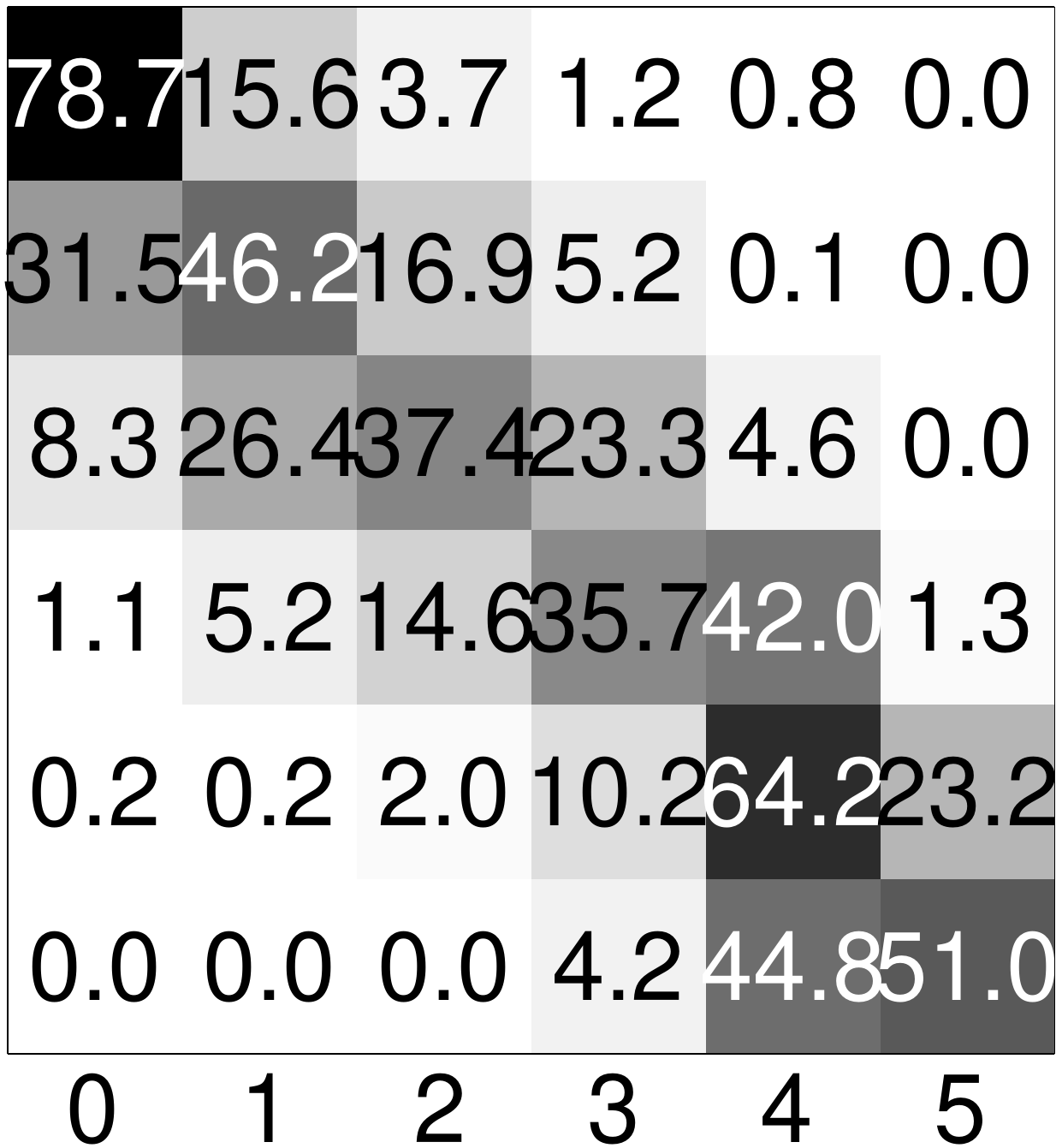}
&\includegraphics[scale=.17]{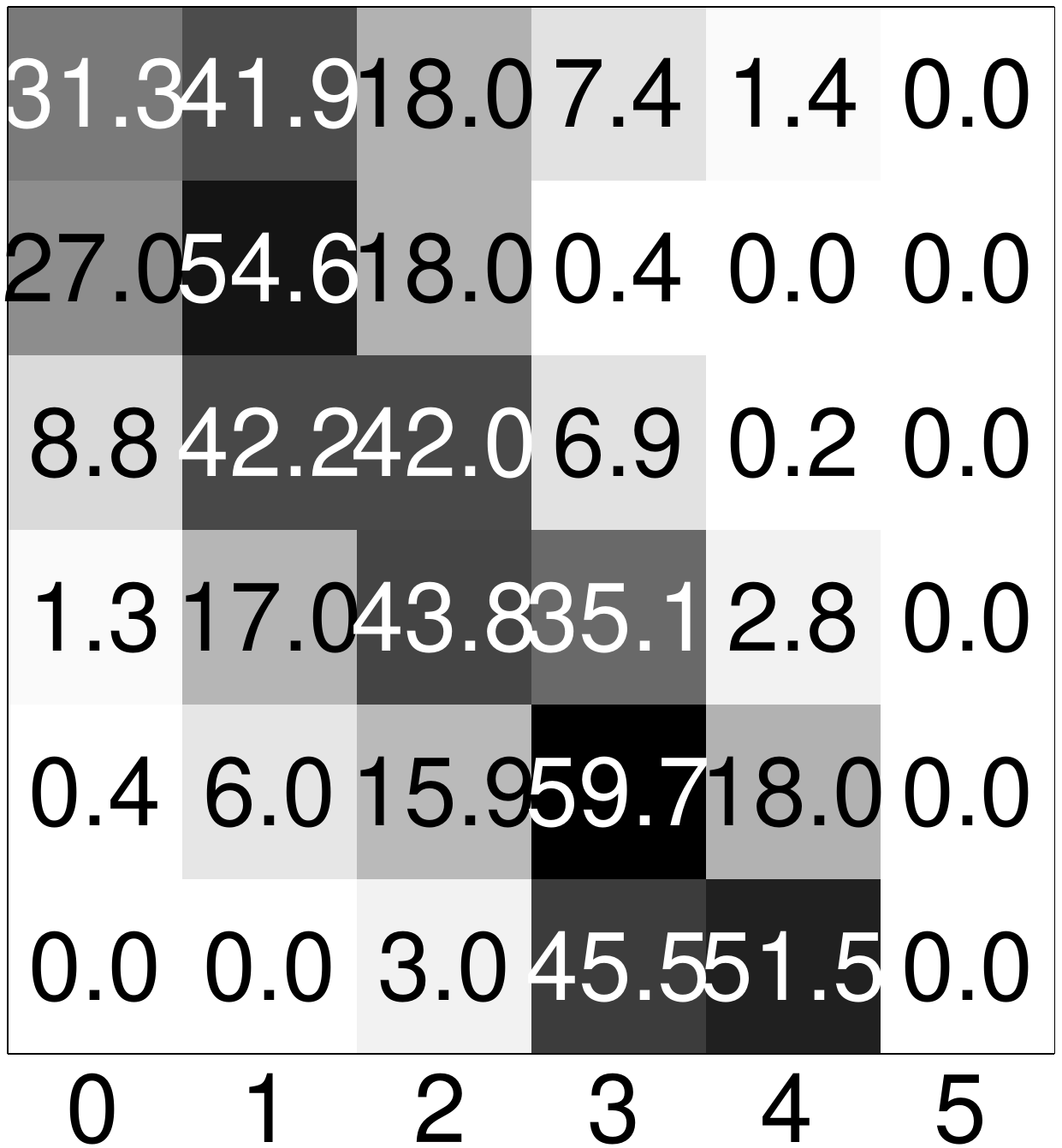}
&\includegraphics[scale=.17]{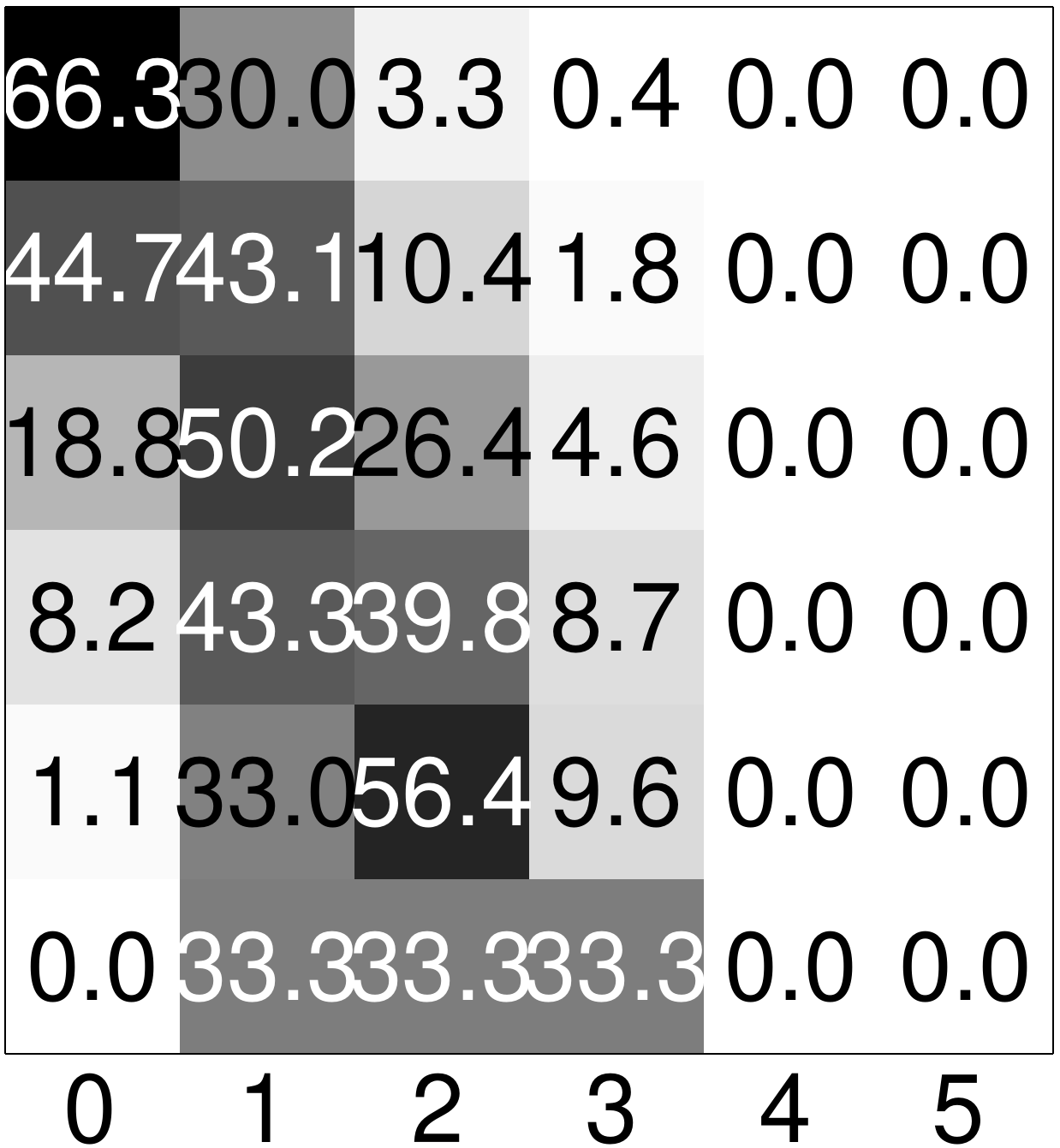}\\
\rot{\rlap{VAE-DGP~\cite{dai2015variational}}} &
\includegraphics[scale=.17]{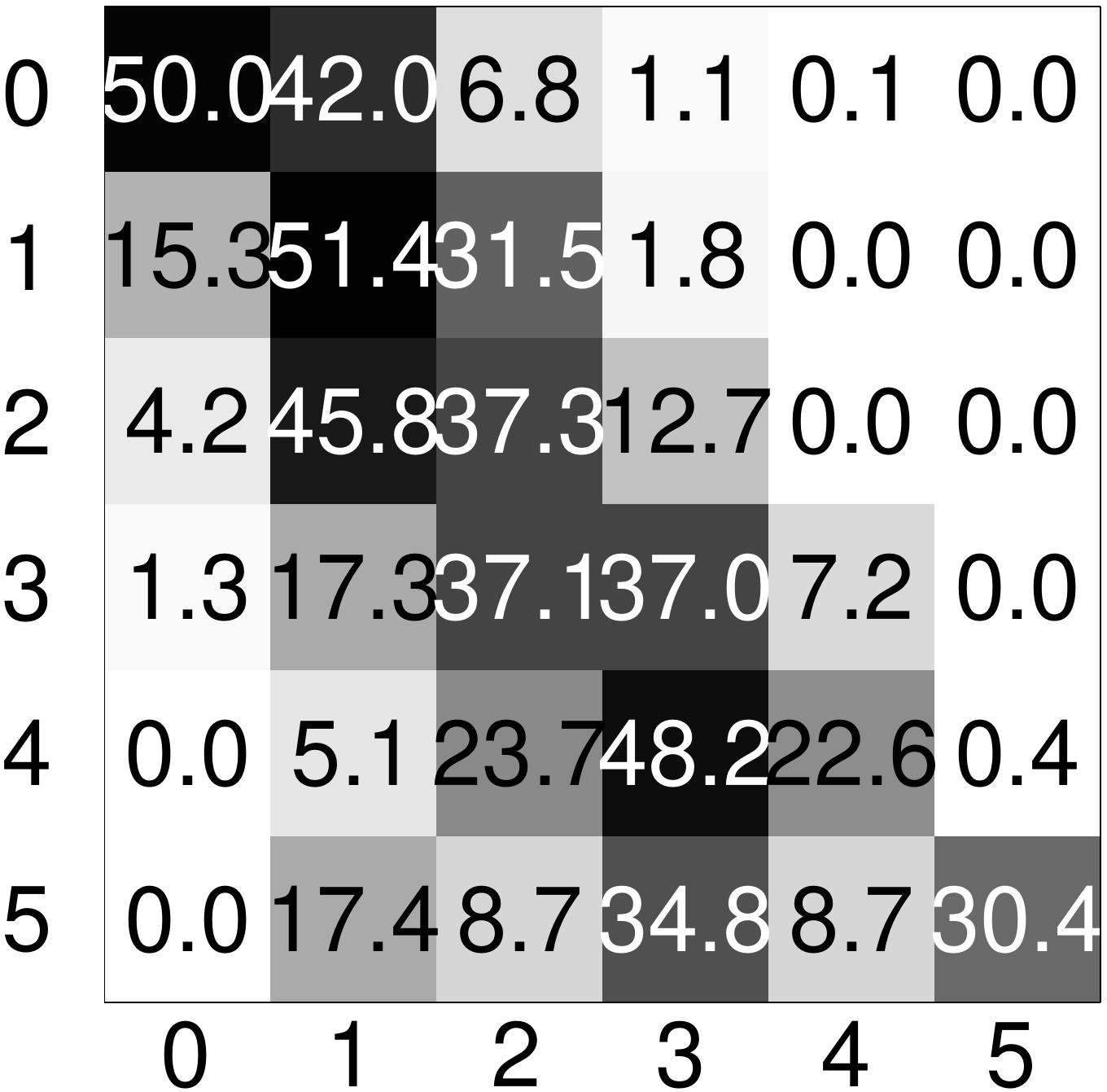} & \includegraphics[scale=.17]{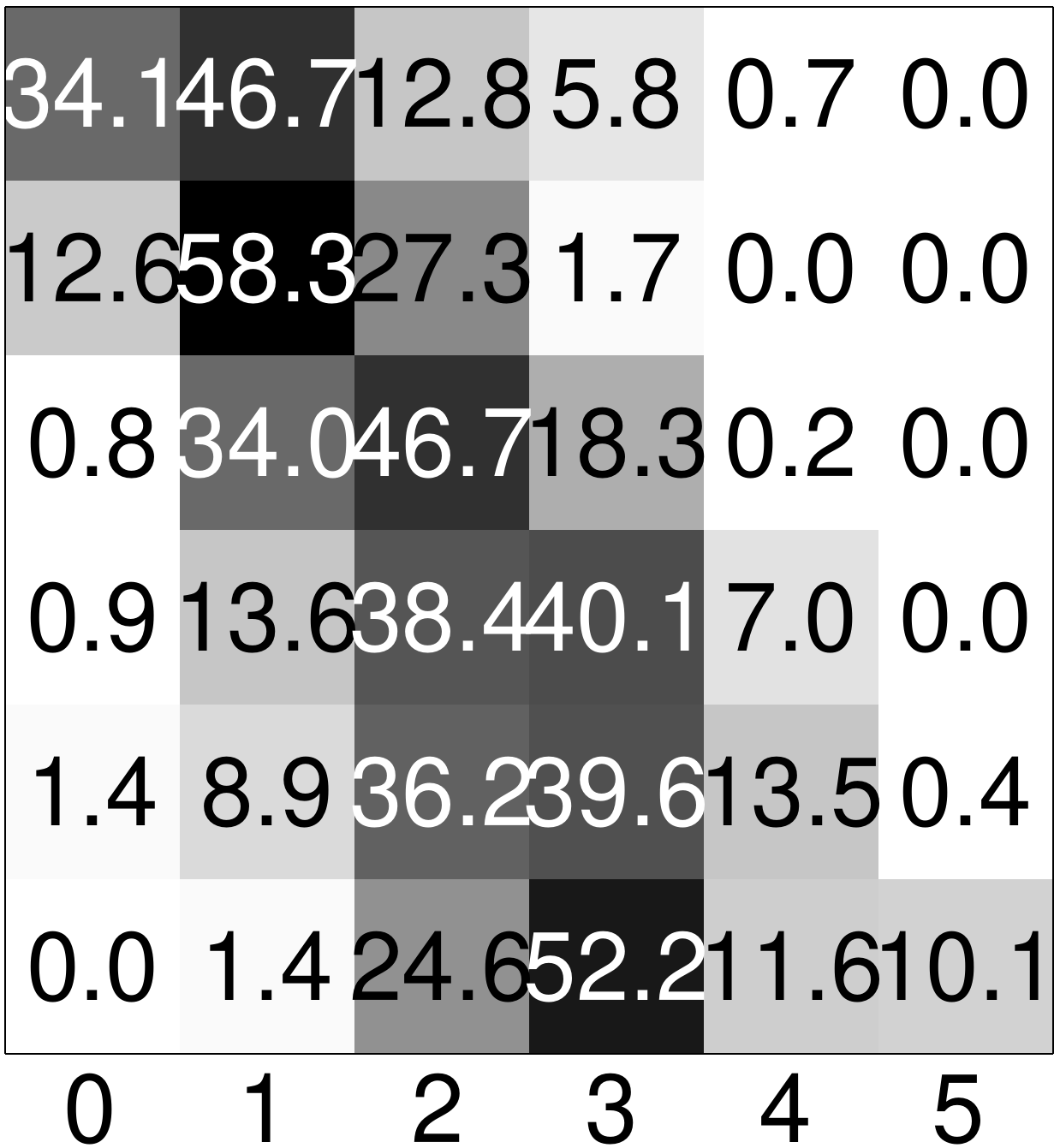}
&\includegraphics[scale=.17]{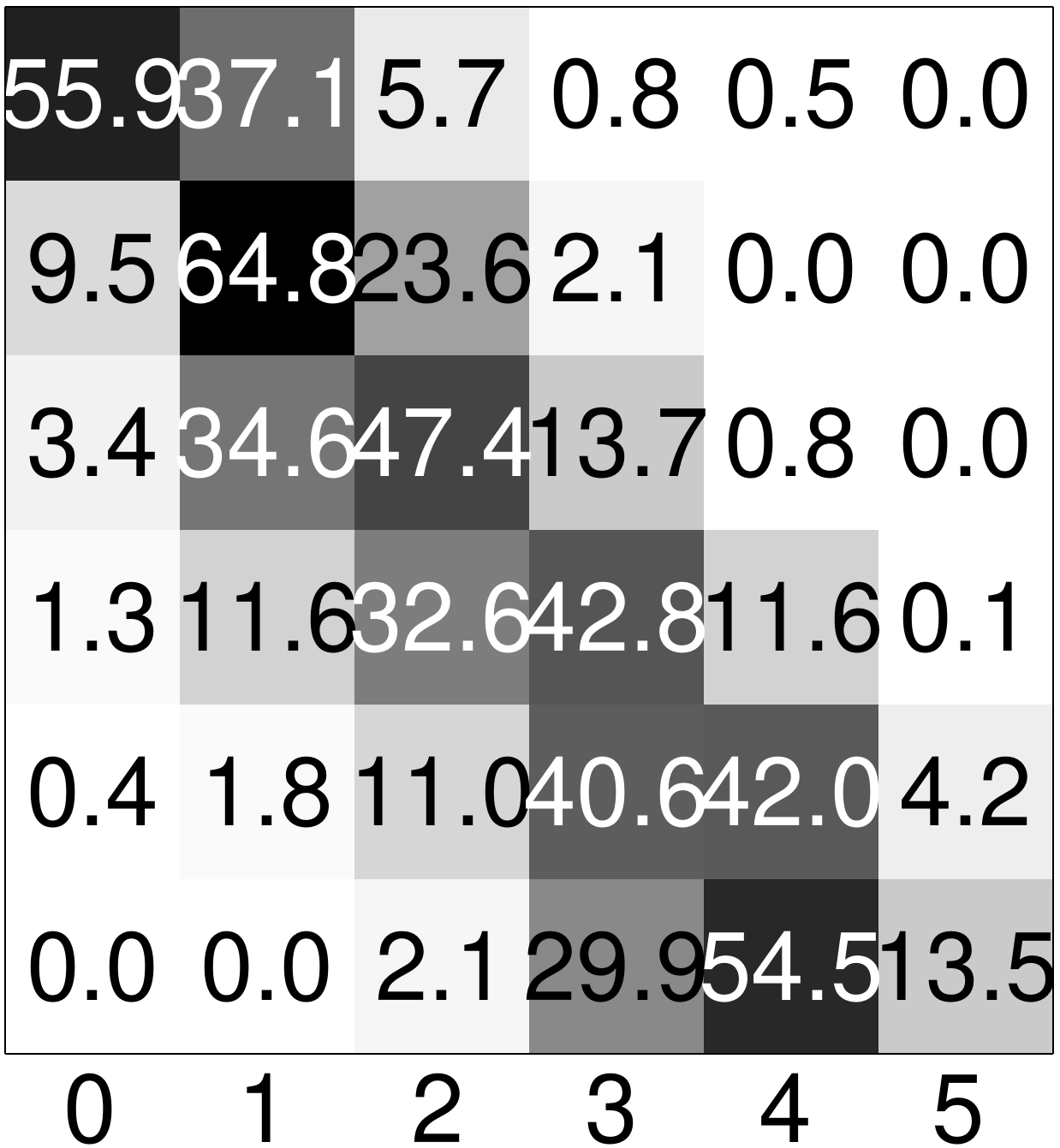}
&\includegraphics[scale=.17]{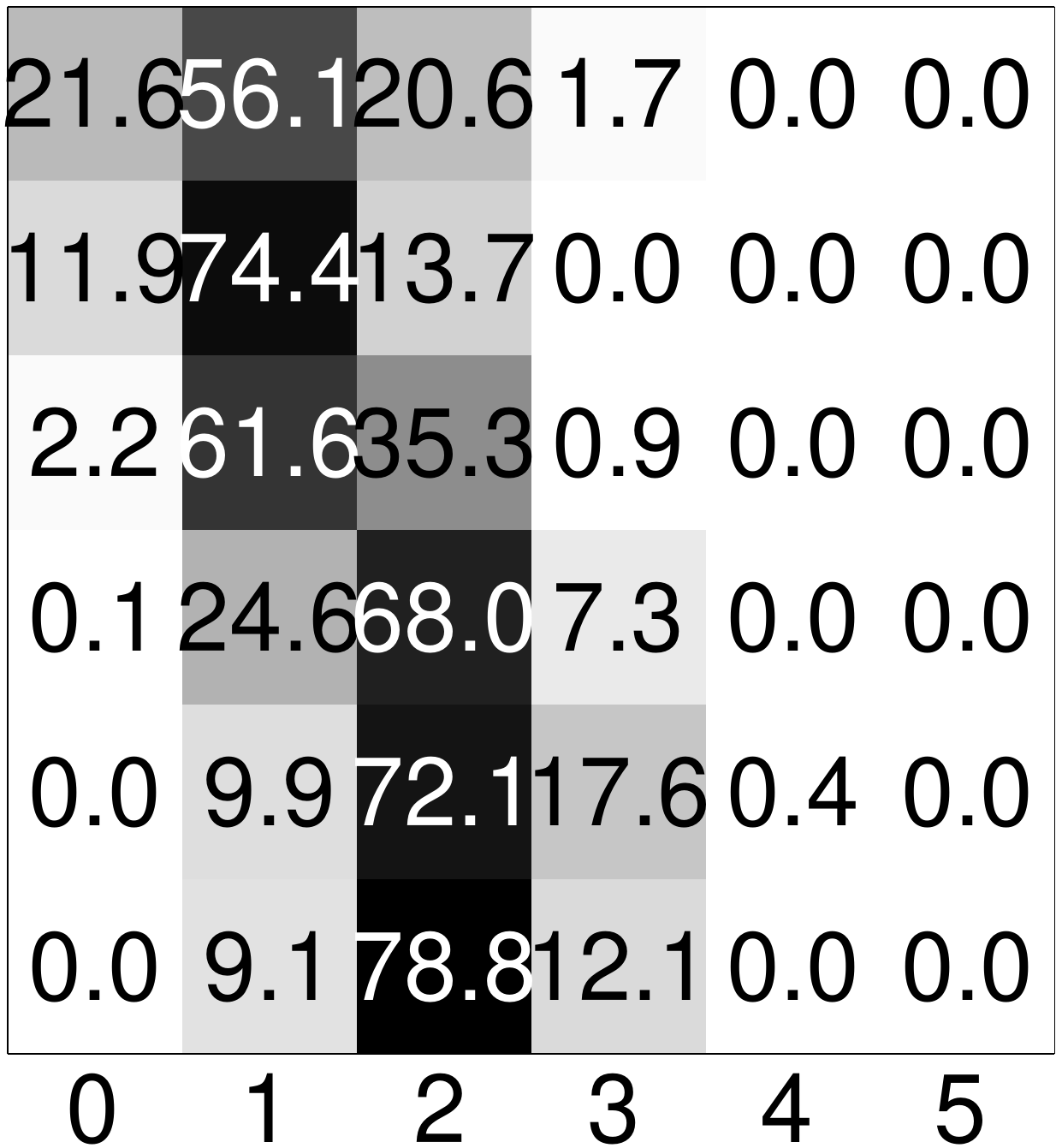}
&\includegraphics[scale=.17]{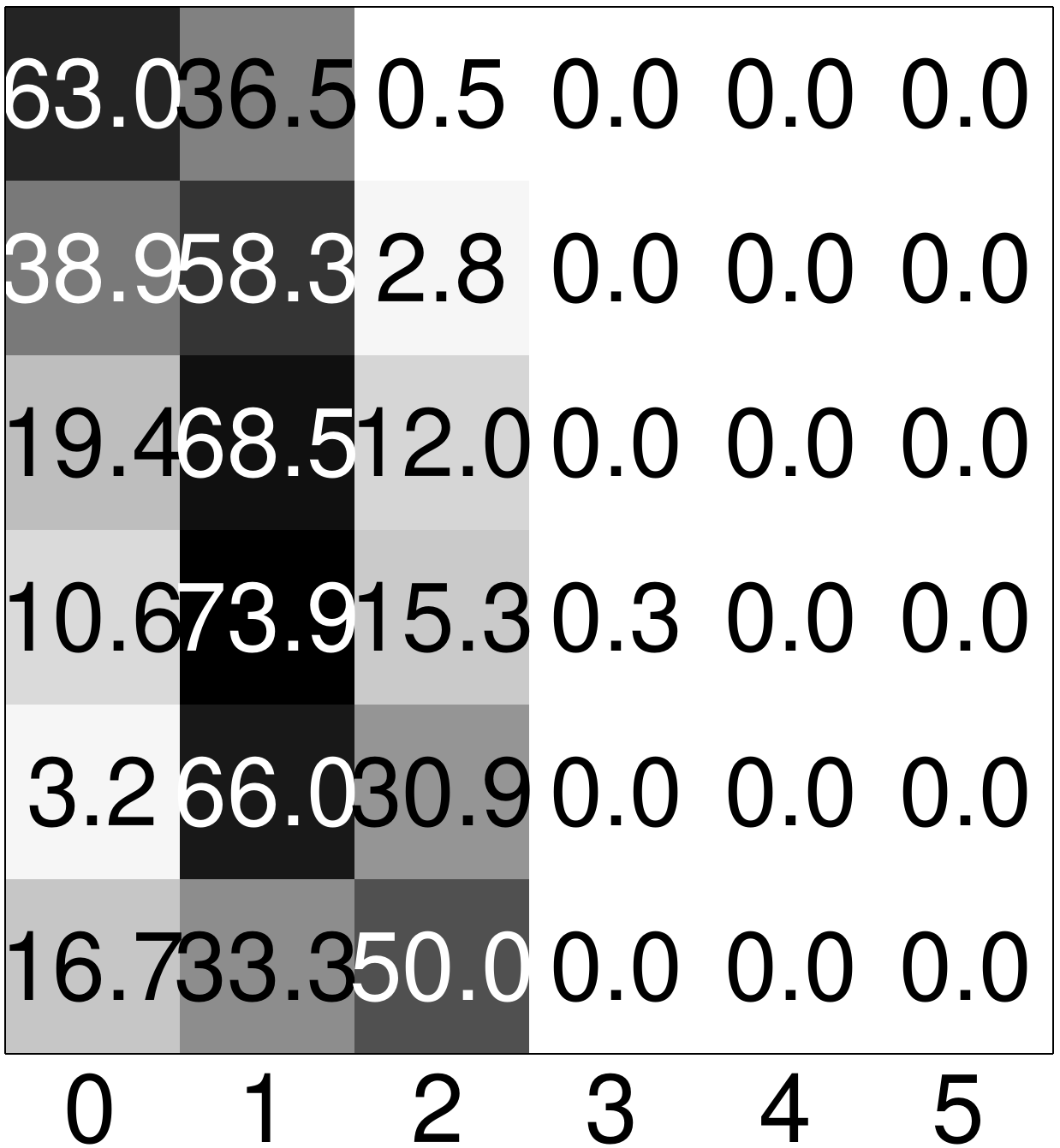}\\
& AU$6$ & AU$10$ & AU$12$ & AU$14$ & AU$17$ 
\end{tabular}
\caption{{\footnotesize Confusion matrices for predicting the $0-5$ intensity of all AUs on FERA2015, when performing fusion with VGP-AE (upper row) and VAE-DGP~\cite{dai2015variational} (lower row).}}
\label{fig_cm}
\end{figure}

The above mentioned difference between our approach and the VAE-DGP is further evidenced in Fig.~\ref{fig_cm}. The proposed fusion along with the novel non-parametric, probabilistic recognition model in our auto-encoder leads to less confusion between the ordinal states across all AUs. We further attribute this to the ordinal modeling of outputs in our VGP-AE, contrary to VAE-DGP that treats the output as continuous variables. This is especially pronounced in the case of the subtle AUs 14\&17, where examples of high intensity levels are scarce.

% \begin{figure}[tb]
% \centering
% \footnotesize
% \setlength{\tabcolsep}{6pt}
% \begin{tabular}{ccccc}
% \rot{\rlap{\qquad \quad AU$6$}} & \includegraphics[scale=.22]{conf_pts_6-cropped.pdf}
% &\includegraphics[scale=.22]{conf_lbp_6-cropped.pdf}
% &\includegraphics[scale=.22]{conf_fusion_6-cropped.pdf}
% &\includegraphics[scale=.22]{conf_fusion_mrd_6-cropped.pdf}\\
% \rot{\rlap{\qquad \quad AU$12$}} & \includegraphics[scale=.22]{conf_pts_12-cropped.pdf}
% &\includegraphics[scale=.22]{conf_lbp_12-cropped.pdf}
% &\includegraphics[scale=.22]{conf_fusion_12-cropped.pdf}
% &\includegraphics[scale=.22]{conf_fusion_mrd_12-cropped.pdf}\\
% & Points & LBP & Fusion & Fusion MRD~\cite{damianou2012manifold}
% \end{tabular}
% \caption{{\footnotesize Confusion matrices for predicting the $0-5$ intensity of AU$6$ (upper row) and AU$12$ (lower row) on FERA2015, when using different modalities (columns). The final column shows the fusion achieved with MRD~\cite{damianou2012manifold}.}}
% \label{fig_cm}
% \end{figure}

%%%%%%%%%%%%%%%%%%%%%%%%%%%%%%%%%%%%%%%%%%%%%%%%%%%%%%%%%%%%%%%%%%%%%%%%%%%%%%%%%
\vspace{-5mm}
\section{Conclusion}
\vspace{-5mm}
We have presented a fully probabilistic auto-encoder, where GP mappings govern both the generative and the recognition models. The proposed variational GP auto-encoder is learned in a supervised manner, where the ordinal nature of the labels is imposed to the manifold. This allows the proposed approach to accurately learn the structure of the input data, while also remain competitive in the classification task. We have empirically evaluated our model on the task of facial feature fusion for joint intensity estimation of facial action units. The proposed model outperforms related GP
methods and the state of the art approaches for the target task.

\vspace{3mm}
%\small
\noindent {\bf Acknowledgement}. This work has been funded by the European Community
Horizon 2020 under grant agreement no. 645094 (SEWA), and no.
688835 (DE-ENIGMA). MPD has been supported by a Google Faculty Research Award.

%===========================================================
\bibliographystyle{splncs}
\bibliography{bibliography}

%\begin{thebibliography}{1}
%
%\bibitem{Alpher02}
%Alpher, A.:
%Advances in Frobnication.
%J. of Foo
%\textbf{12} (2002)  234--778
%
%\bibitem{Alpher03}
%Alpher, A., Fotheringham-Smythe, J.P.N.:
%Frobnication revisited.
%J. of Foo
%\textbf{13} (2003)  234--778
%
%\bibitem{Herman04}
%Herman, S., Fotheringham-Smythe, J.P.N., Gamow, G.:
%Can a machine frobnicate?
%J. of Foo
%\textbf{14} (2004)  234--778
%
%\bibitem{Smith09}
%Smith, F.:
%{\it The Frobnicatable Foo Filter}.
%GreatBooks, Atown (2009)
%
%\bibitem{Wills99}
%Wills, H.:
%Frobnication tutorial.
%Technical report CS-1204, XYZ University, Btown (1999)
%
%\end{thebibliography}

%this would normally be the end of your paper, but you may also have an appendix
%within the given limit of number of pages
%\end{document}

%===========================================================

\end{document}